\documentclass[letterpaper]{article} 
\usepackage[preprint]{aaai2027}  
\usepackage[hyphens]{url}  
\usepackage{graphicx} 
\usepackage{natbib}  
\usepackage{caption} 
\usepackage{booktabs}

\usepackage{array}
\newcolumntype{C}[1]{>{\centering\arraybackslash}p{#1}}
\newcolumntype{L}[1]{>{\raggedright\arraybackslash}p{#1}}

\usepackage{longtable}
\usepackage{multirow}
\usepackage{placeins} 

\title{GISAgentBench: A Practitioner-Sourced Benchmark for Evaluating LLM Agents on GIS Tasks}
\author{
    Abhinav Pothuri\textsuperscript{\rm 1},
    Zhe Jiang\textsuperscript{\rm 1}\thanks{Contact author: Zhe Jiang \texttt{zhe.jiang@ufl.edu}},
    Zelin Xu\textsuperscript{\rm 1},
    Di Yang\textsuperscript{\rm 2}
}
\affiliations{
    \textsuperscript{\rm 1}Department of Computer \& Information Science \& Engineering, University of Florida\\
    \textsuperscript{\rm 2}Department of Geography, University of Florida
}

\begin{document}

\maketitle

\begin{abstract}
Geographic Information System (GIS) professionals rely on multi-step spatial analysis workflows to support decision-making in urban planning, disaster response, and environmental monitoring. The process is tedious, time-consuming, and error-prone. While recent large language model (LLM) agents equipped with external tools have the potential to automate geospatial analysis, their ability to perform realistic GIS workflows remains largely unexplored. Existing GIS agent benchmarking datasets are mostly drawn from textbooks, tutorials, or LLM-generated seeds, and remain limited in size and trajectory depth. More importantly, none provides ground truth outputs. They therefore rely on surrogate signals such as code similarity, trajectory matching, or LLM and VLM judges, which can conflate workflow resemblance with task correctness. To address this gap, we introduce \textbf{GISAgentBench}, a benchmark of 349 multi-step GIS tasks curated from GIS Stack Exchange and instantiated on real public data across six selected geographic areas of interest. Each task ships with an executable reference trajectory and an exact ground truth output file, enabling strict, deterministic, tolerance-aware output matching beyond LLM judging. Evaluations of six LLM models reveal that realistic GIS workflows remain challenging: the best agent completes only 32.7\% of tasks under strict tolerance-aware scoring, although most models produce outputs that are close to the ground truth. 
\end{abstract}


\section{Introduction}
\label{sec:intro}

A geographic information system (GIS) is a software system for storing, managing, analyzing, and visualizing spatially referenced data. It plays a crucial role in addressing grand societal challenges such as disaster management, urban planning, land administration, climate adaptation, and precision agriculture~\cite{goodchild2010gis}. For example, the Federal Emergency Management Agency delineates flood hazard zones by analyzing terrain, hydrography, and parcel data in GIS, producing the maps that underpin the National Flood Insurance Program and its \$1.3 trillion of coverage~\cite{fema}. Yet such products are rarely the result of a single operation. Even a well-defined task, such as identifying census tracts within a floodplain and computing their demographic overlap, requires an analyst to gather heterogeneous datasets, reconcile their coordinate systems and schemas, select and parameterize operations from toolboxes and libraries of hundreds of primitives, and execute a long chain of geoprocessing steps. Today, such analyses are performed daily by a large professional workforce of GIS analysts and technicians, working either through graphical map interfaces or scripted geoprocessing pipelines~\cite{bls}. Unfortunately, professional GIS work remains tedious, time-consuming, and error-prone.  Recently, large language models (LLMs) capable of tool use~\cite{toolformer} and multi-step reasoning~\cite{react,cot} have provided a promising alternative that could drive a major paradigm shift. Given a natural-language request, an LLM agent can in principle decompose the task into spatial operations, invoke GIS tools to execute the workflow, and return the final spatial product. Early systems that generate and execute GIS code or spatial queries demonstrate this potential~\cite{llmgeo,geogpt,geoagent}. An agent that performed such workflows reliably would both automate routine geospatial labor and make advanced spatial analysis accessible to non-specialists.

\begin{table*}[t]
\centering
\caption{Comparison of existing GIS agent benchmarks.}
\label{tab:comparison}
\small
\setlength{\tabcolsep}{4pt}
\renewcommand{\arraystretch}{1.3}
\begin{tabular}{l c L{2.8cm} c C{2.7cm} L{3.2cm} L{2.2cm}}
\toprule
\textbf{Benchmark} & \textbf{\# Tasks} & \textbf{Source} & \textbf{\# Tools} & \textbf{Traj.\ Length} & \textbf{Evaluation Method} & \textbf{Challenge Tags} \\
\midrule
GeoAnalystBench & 50  & Textbook \& Tutorials                   & unknown & Avg 6; Max 10  & CodeBLEU to match code          & --        \\
GeoBenchX       & 202 & LLM-generated from a few industry seeds & 23  & Avg 5; Max 14  & LLM judge to match trajectory    & Post hoc error analysis \\
GeoAgentBench   & 53  & Textbook \& Tutorials                   & 117 & Avg 7; Max 17  & Trajectory match + VLM judge on map appearance & --        \\
\midrule
\textbf{GISAgentBench} & \textbf{349} & \textbf{GIS Stack Exchange} & \textbf{128} & \textbf{Avg 12; Max 43} & \textbf{Tolerance-aware output matching + traj. match} & \textbf{Pre-identified human pitfalls} \\
\bottomrule
\end{tabular}
\end{table*}

However, GIS tasks are uniquely challenging for LLM agents. Professional GIS work is inherently cross-disciplinary, drawing simultaneously on geospatial information science, spatial databases, GIS software, machine learning, statistics, and remote sensing. This interdisciplinary character produces failure modes largely absent from general-purpose code generation and agent benchmarks. First, an undetected coordinate reference system (CRS) mismatch silently invalidates every distance, area, and intersection computed downstream. Second, practitioner requests are often underspecified, leaving the intended spatial predicate, measurement unit, or attribute schema to be inferred. A request for features \emph{inside} a boundary, for example, does not say whether features touching that boundary should be retained. Third, real spatial data are imperfect. Rasters carry NoData masks, and vector layers contain invalid, overlapping, or multipart geometries. Fourth, GIS analyses are long chains of interdependent operations, so an error introduced early propagates through every subsequent step. A workflow can therefore run to completion without a single error and still be spatially wrong. Finally, evaluation is itself difficult, because correct outputs may legitimately differ in geometric representation or fall within numerical tolerances. This raises a fundamental open question: \emph{how reliably can frontier LLM agents solve the multi-step GIS tasks that practitioners face in real-world settings?}

Existing benchmarks cannot answer this question. Systems that apply LLMs to GIS code generation, spatial database querying, urban analytics, and map understanding~\cite{llmgeo,geogpt,urbangpt,citygpt} are largely demonstrations on small handcrafted examples rather than standardized evaluations. General agent and code benchmarks~\cite{agentbench,gaia,webshop,toolbench,humaneval,ds1000,swebench} define correctness over text answers, web interactions, or unit tests, none of which capture spatial outputs, and evaluations of geospatial code generation~\cite{gramacki2024geospatial,geocodegpt} measure code synthesis rather than the full loop of planning, tool selection, geoprocessing, and output validation. Spatial reasoning benchmarks~\cite{geoqa,geosqa,spatialvlm,earthspatialbench} assess geographic knowledge or map interpretation without tool-mediated geoprocessing. The closest works are GeoAnalystBench~\cite{geoanalystbench}, GeoBenchX~\cite{geobenchx}, and GeoAgentBench~\cite{geoagentbench}, which evaluate tool-calling agents on multi-step geospatial tasks (Table~\ref{tab:comparison}). However, their tasks are drawn from textbooks, tutorials, or LLM-generated seeds rather than from practice, and remain limited in size (50 to 202 tasks) and depth (reference trajectories of 5 to 7 tool calls on average, at most 17). More importantly, none of them provides an executable ground truth output. They therefore rely on surrogate signals such as code similarity, trajectory matching, or LLM and VLM judges, which can conflate workflow resemblance with task correctness. In addition, none of them identifies in advance where a task is prone to error.

To fill this gap, we introduce \textbf{GISAgentBench}, a benchmark of 349 multi-step GIS tasks curated from GIS Stack Exchange~\cite{gisstackexchange}, the largest public question-and-answer forum for GIS practitioners, and instantiated on real public datasets across six geographic areas of interest. To our knowledge, it is the first GIS agent benchmark built from community-generated practitioner questions and scored against exact ground truth outputs. Practitioner-based sourcing gives our benchmark five distinguishing features. First, the tasks are real-world problems posed by working analysts, rather than textbook or tutorial exercises. Second, each task is paired with an executable reference solution and an exact ground truth output file, so that agents are scored by strict, deterministic, tolerance-aware output matching. Third, we also evaluate agent trajectories alongside final outputs, which separates planning failures from execution failures. Fourth, our benchmark is substantially larger and deeper than related work, with 349 tasks whose reference solutions require up to 43 API calls (11.7 on average), compared with around 100 tasks and far shorter trajectories in prior work. Finally, and most importantly, we annotate each task in advance with practical caveats, namely the CRS, unit, NoData, boundary, and geometry pitfalls drawn from the practical experience of GIS professionals. These conditions are highly error-prone even for human experts, and annotating them allows us to evaluate whether LLM agents fall into the same traps as human practitioners in failture mode analysis. Using this benchmark, we evaluate six LLM agents under a standardized framework with a fixed harness of 128 GIS APIs, deterministic tolerance-aware output matching, and trajectory-level diagnostics. The results show that realistic GIS workflows remain far beyond the current capability frontier, with the best agent solving only 32.7\% of tasks under strict scoring. A quantitative closeness analysis further reveals that agents recover geometry far more reliably than numeric attributes and complete row sets, and that our pre-annotated pitfalls are predictive of failure, with geometry and topology errors and CRS misalignment the most damaging. The complete task set, reference trajectories, and ground truth outputs will be released publicly.

\section{GIS Agent Benchmark}
\label{sec:dataset}

\begin{figure*}[t]
\centering
\includegraphics[width=\textwidth, keepaspectratio]{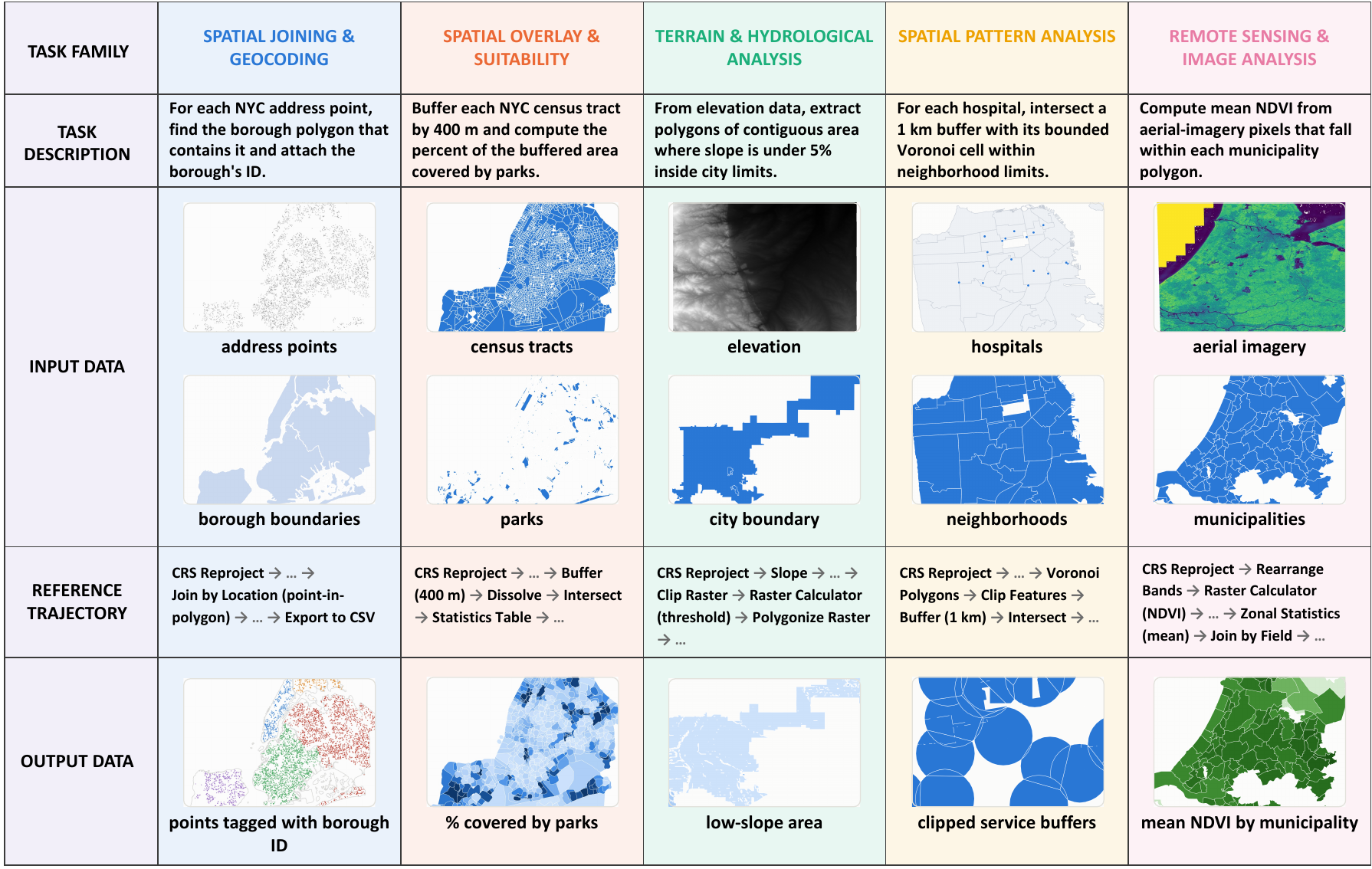}
\caption{Illustrative examples of GISAgentBench tasks.}
\label{fig:task_examples}
\end{figure*}

GISAgentBench is a benchmark of \textbf{349 multi-step GIS tasks} sourced from practitioner questions on GIS Stack Exchange and instantiated on real public spatial data across six geographic areas of interest. Each task pairs a practitioner-style problem statement with its input datasets, an output contract specifying the required file and how it will be scored, an executable reference trajectory, and expert annotations recording which task family the problem belongs to and which practical pitfalls it exercises.

Tasks are organized into five families:
\begin{itemize}
\item \textbf{Spatial Joining \& Geocoding (SJ\&G)}: associating records, coordinates, or features across spatial and tabular datasets (96 tasks).
\item \textbf{Spatial Overlay \& Suitability Analysis (SO\&SA)}: combining layers through predicates, buffers, intersections, or selection criteria to identify areas satisfying spatial constraints (93 tasks).
\item \textbf{Terrain Modeling \& Hydrological Analysis (TM\&HA)}: deriving products from elevation data, such as slope, drainage, watershed, and flow-related outputs (63 tasks).
\item \textbf{Spatial Pattern Analysis (SPA)}: measuring geometric, topological, or distributional structure, including clustering, proximity patterns, thinning, and morphology (52 tasks).
\item \textbf{Remote Sensing \& Image Analysis (RS\&IA)}: processing raster imagery or derived indices through masking, classification, pixel-level computation, or zonal summarization (45 tasks).
\end{itemize}

Figure~\ref{fig:task_examples} shows one example from each family, which makes the shape of a task concrete. An agent receives a request phrased as an analyst would pose it, together with a set of data files, and must return a specific file whose columns, coordinate system, and numeric tolerances are fixed in advance. Solving it means composing named geoprocessing operations, each of which reads one or more \emph{layers}---tables whose rows carry a geometry---and writes a new one. Some operations relate layers by position, such as finding which points fall inside which polygons. Others construct new geometry, such as the zone within a fixed distance of a feature. Others summarize a grid of pixels into one value per region. Because each call consumes the file the previous call produced, a task is a dependent chain rather than a set of independent queries, and what is scored is a spatial file rather than a string.

\noindent\textbf{GIS API harness:} Agents interact with spatial data exclusively through a fixed harness of \textbf{128 GIS APIs}, backed by QGIS through its command-line interface, including its GDAL-backed algorithms, together with GeoPandas, Rasterio, Shapely, pandas/NumPy, and SciPy. Every API is exposed through a uniform interface with typed arguments and a documented return contract, and the set is identical across all evaluated models. Table~\ref{tab:api_categories} summarizes its functional coverage. Appendix~C lists all 128 with descriptions and parameters. This fixed-harness design is central to our evaluation: every task is verified to be solvable with the provided APIs, so tool availability is never a confounding factor; and holding the harness constant isolates an agent's ability to select, parameterize, and sequence GIS operations from differences in what harness it can reach.

\begin{table}[t]
\centering
\small
\setlength{\tabcolsep}{4pt}
\caption{Functional coverage of the 128-API harness.}
\label{tab:api_categories}
\begin{tabular}{lr@{\hskip 1.2em}lr}
\toprule
\textbf{Category} & \textbf{\#} & \textbf{Category} & \textbf{\#} \\
\midrule
Geometry construction   & 22 & Spatial join          & 6 \\
Attribute \& table ops   & 15 & Inspection            & 6 \\
Raster processing       & 15 & Zonal statistics      & 5 \\
Raster--vector          & 11 & CRS \& reprojection    & 4 \\
Buffers \& proximity     & 10 & Network analysis      & 4 \\
Terrain \& hydrology     &  9 & Density surfaces      & 3 \\
Vector overlay          &  8 & Input/output          & 3 \\
Geometry cleaning       &  7 & \textbf{Total}        & \textbf{128} \\
\bottomrule
\end{tabular}
\end{table}

\subsection{Task Set Selection and Construction}
\label{sec:schema}

\begin{figure}[t]
\centering
\includegraphics[width=\columnwidth]{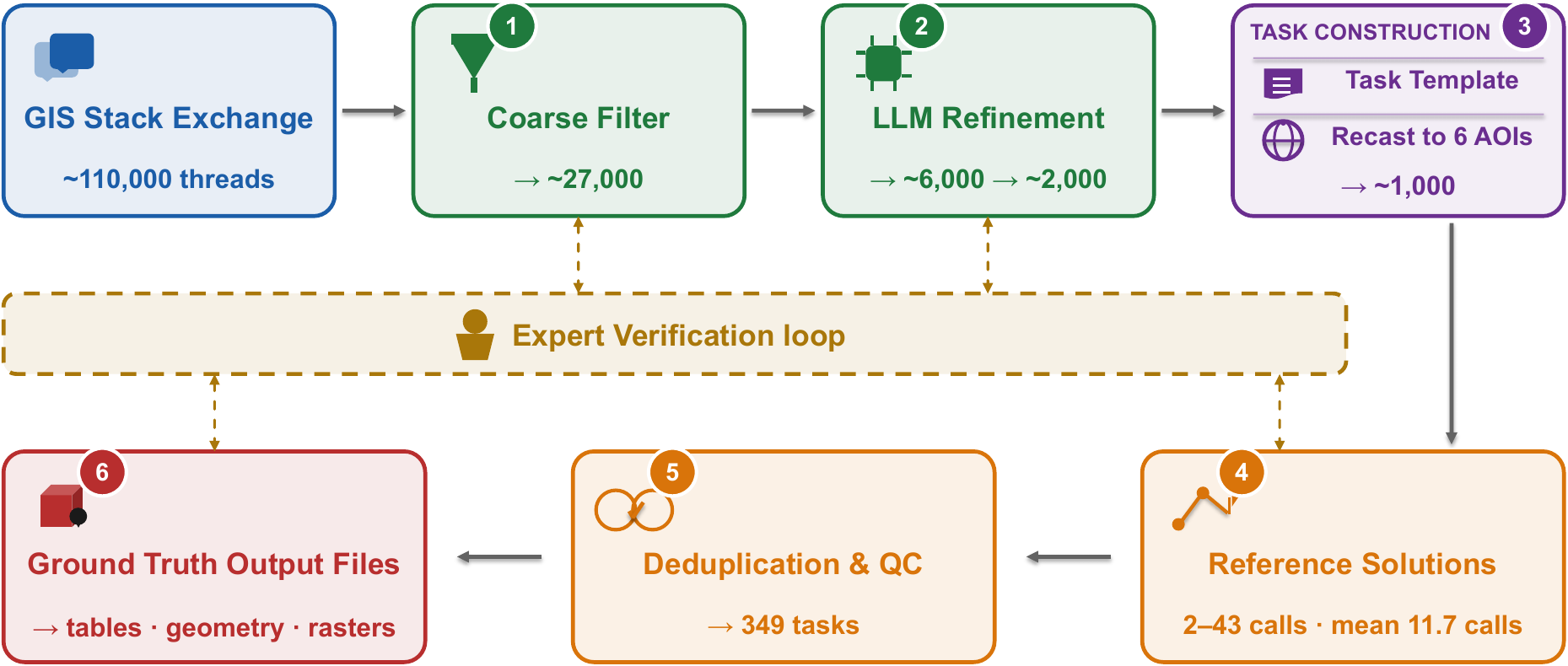}
\caption{GISAgentBench construction pipeline.}
\label{fig:dataset_construction}
\end{figure}

GISAgentBench is built from GIS Stack Exchange, the largest public question-and-answer forum for GIS practitioners. Unlike benchmarks that rely on textbook, tutorial, or synthetic tasks, it is grounded in problems encountered during operational geospatial analysis: questions are posed organically by working analysts, reflecting authentic analytical needs and workflow complexity, while community-vetted answers provide a foundation for constructing reference solutions.

Figure~\ref{fig:dataset_construction} summarizes the construction pipeline. From approximately 110{,}000 crawled threads, a deterministic filter retains roughly 27{,}000 by scoring each thread on GIS terminology, workflow verbs, and data context, and rejecting posts that are too short, that concern software setup or map display rather than spatial analysis, or that repeat an earlier post. Two LLM screening passes with GPT-5-mini follow: a lightweight triage discards threads without a concrete spatial goal, leaving about 6{,}000, and a stricter pass retains only problems that require more than a single operation, are reproducible from fixed inputs, and can be moved to new data without altering the underlying question, leaving roughly 2{,}000 candidates. Both screening prompts were developed iteratively rather than written once: a GIS expert reviewed initial batches of 50--100 threads against each draft of the criteria, and recurring errors were folded back into the prompts until the retained set matched expert judgment. 
Each retained thread is then cast into a common benchmark task template: a practitioner-style prompt, the input data in a selected area of interest, the task parameters, and an output contract naming the required file, columns, spatial reference, and tolerances. After recasting the task description into a fixed set of areas of interest (Section~\ref{sec:aoi}) and generating reference trajectories (Section~\ref{sec:gt}), we remove tasks whose reference cannot be produced from the declared inputs, whose trajectories are duplicated, or whose trajectories prove too short to require multi-step reasoning. We finally collapse tasks that reduce to the same ordered sequence of operations, leaving a final set of \textbf{349 tasks}. \textbf{Appendix~B} gives the complete keyword dictionaries, filter thresholds, and screening and drafting prompts.

Each task is also tagged by GIS experts with the \emph{caveats} it exercises: CRS misalignment (244 tasks), boundary ambiguity (180), geometry and topology errors (113), NoData values (90), and unit mismatch (47). Annotating pitfalls in advance lets us ask whether LLM agents fall into the same traps as human experts. Table~\ref{tab:crosstab} cross-tabulates the two annotations, and the caveat profile follows the data model rather than being uniform. The vector families are dominated by coordinate-system and geometry problems: 90\% of Spatial Joining \& Geocoding tasks carry a CRS caveat and 46\% a geometry one, while NoData is almost absent from them (3 of 96). The raster families invert this, with NoData in 59\% of Terrain Modeling and 62\% of Remote Sensing tasks.

\begin{table}[t]
\centering\small
\setlength{\tabcolsep}{3pt}
\caption{Task family by practitioner caveat.}
\label{tab:crosstab}
\begin{tabular}{@{}lcccccrr@{}}
\toprule
 & \textbf{CRS} & \textbf{Bound.} & \textbf{Geom.} & \textbf{NoData} & \textbf{Unit} & \textbf{none} & \textbf{Tasks} \\
\midrule
SJ\&G  &  86 &  68 & 44 &  3 & 13 &  1 &  96 \\
SO\&SA &  71 &  47 & 30 & 12 & 19 &  4 &  93 \\
SPA    &  31 &  21 & 20 & 10 &  8 &  3 &  52 \\
RS\&IA &  16 &  22 &  6 & 28 &  3 &  4 &  45 \\
TM\&HA &  40 &  22 & 13 & 37 &  4 &  4 &  63 \\
\midrule
\textbf{Total} & \textbf{244} & \textbf{180} & \textbf{113} & \textbf{90} & \textbf{47} & \textbf{16} & \textbf{349} \\
\bottomrule
\end{tabular}
\end{table}

\subsection{Geographic Recasting into Selected Areas of Interest}
\label{sec:aoi}

Because the source threads are public, their accepted answers are indexed on the web and may sit in any model's pretraining data. Geographic recasting mitigates that leakage: an agent receives the problem posed over data the original thread never used, so no published answer exists for it. Schema normalization on import extends the same protection to field names. In this process, we preserve a task's analytical objective, spatial predicates, and data roles while changing the geography, datasets, schemas, and resulting numbers.

This demands target geographies that are individually rich and collectively diverse. The six areas of interest (Table~\ref{tab:aoi}) were chosen to span the task families rather than repeat one layer stack: dense urban-coastal and cadastral settings for joins, overlays, and topology repair; a mountain front and a low-relief floodplain for terrain, hydrology, and least-cost analysis; an engineered water-control landscape for constrained routing; and a nationally standardized European environment that tests the same operations outside U.S. data conventions. Each area draws on public national, municipal, and European open-data portals at operational scale, spanning vector, raster, and tabular layers. To ensure that a task retains the caveat (pitfall) tagged in its source GIS Stack Exchange post, a few input datasets are slightly modified. Appendix~D lists every data source, the principal layers with exact feature counts, and every raster with its full characteristics.

\begin{table}[t]
\centering
\small
\setlength{\tabcolsep}{4pt}
\caption{The six areas of interest. Appendix~D details their datasets and per-family task counts.}
\label{tab:aoi}
\begin{tabular}{@{}L{2.35cm} L{4.35cm} r@{}}
\toprule
\textbf{Area of interest} & \textbf{Landscape} & \textbf{Tasks} \\
\midrule
San Francisco Bay Area & Dense urban coastal; bay shoreline, terrain, imagery & 99 \\
Randstad, Netherlands  & Low-lying delta; nationally standardized registries  & 91 \\
New York City          & Dense urban; parcel- and address-rich cadastre       & 51 \\
Denver Front Range     & Mountain front; wildland--urban interface, hazards   & 40 \\
Houston--Harris        & Flat coastal plain; floodplain and drainage network  & 40 \\
South Florida          & Engineered water control; low-relief coastal wetland & 28 \\
\bottomrule
\end{tabular}
\end{table}

\begin{table*}[!t]
\centering
\caption{Model performance on tasks measured in strict TSR and quantitative closeness, overall and by task family.}
\label{tab:overall_results}
\small
\setlength{\tabcolsep}{3.5pt}
\begin{tabular}{lcccccc cccccc}
\toprule
& \multicolumn{6}{c}{\textbf{Strict Task Success Rate (TSR)}} & \multicolumn{6}{c}{\textbf{Quantitative Closeness Score (QCS)}} \\
\cmidrule(lr){2-7} \cmidrule(lr){8-13}
\textbf{Model} & \textbf{All} & \textbf{SJ\&G} & \textbf{SO\&SA} & \textbf{SPA} & \textbf{RS\&IA} & \textbf{TM\&HA} & \textbf{All} & \textbf{SJ\&G} & \textbf{SO\&SA} & \textbf{SPA} & \textbf{RS\&IA} & \textbf{TM\&HA} \\
\midrule
Gemini-3.1-Pro   & \textbf{0.327} & \textbf{0.385} & \textbf{0.383} & \textbf{0.231} & \textbf{0.378} & 0.190 & \textbf{0.663} & \textbf{0.676} & \textbf{0.680} & \textbf{0.623} & \textbf{0.703} & 0.628 \\
Claude-4.6-Opus  & 0.295 & 0.302 & 0.344 & \textbf{0.231} & 0.333 & 0.238 & 0.606 & 0.569 & 0.640 & 0.570 & 0.546 & \textbf{0.688} \\
DeepSeek-V4-Pro  & 0.275 & 0.312 & 0.277 & 0.212 & 0.244 & \textbf{0.286} & 0.595 & 0.571 & 0.620 & 0.483 & 0.636 & 0.658 \\
GPT-5.4          & 0.229 & 0.188 & 0.309 & 0.173 & 0.200 & 0.238 & 0.538 & 0.515 & 0.606 & 0.381 & 0.524 & 0.619 \\
Qwen3.6-27B      & 0.178 & 0.188 & 0.213 & 0.135 & 0.156 & 0.159 & 0.436 & 0.429 & 0.484 & 0.360 & 0.483 & 0.414 \\
GPT-OSS-120B     & 0.123 & 0.125 & 0.138 & 0.115 & 0.133 & 0.095 & 0.380 & 0.365 & 0.377 & 0.346 & 0.361 & 0.437 \\
\bottomrule
\end{tabular}
\end{table*}

\subsection{Ground Truth Reference Trajectory and Output Generation}
\label{sec:gt}

Each task ships with an executable reference trajectory and the output file it produces, which are generated and verified by the Claude Fable 5 model (a current frontier model that is deliberately not among our evaluated agents). The released trajectory is expressed entirely in the same 128-API harness the agents are given, and is executed inside that harness to materialize the ground truth output file. In order to cross-check these outputs, we also provide Fable 5 direct access to the input data and to free-form Python for deriving and checking what the correct result should be. Across the benchmark these trajectories range from 2 to 43 API calls (mean 11.7) over 92 distinct input files, reflecting the compositional depth of real practitioner workflows.

\textbf{Quality assurance} proceeds in two steps. First, an automated recheck statically validates every reference step against the real API signatures, catching fake tools, misspelled arguments, and unbound intermediate outputs. Second, and more importantly, three GIS experts (in our author list) reviewed the reference trajectories directly. Trajectories were not accepted merely because a model produced them or because a command sequence ran to completion: they were hand-traced step by step, with manual inspection of intermediate outputs and visual checks of the resulting spatial outputs. Expert review is what surfaced the defect classes that automated checks cannot see, such as a field calculation bound to a column the upstream operation never produced, a slope computed on an unprojected elevation model, or an output contract keyed on a column the prompt never promised. Tasks whose descriptions proved ambiguous, or whose specification required revision to be solvable, were removed rather than repaired in place. Appendix~E documents the details.

\section{Experimental Evaluation}
\label{sec:framework}

\subsection{Experimental Setup}

\noindent\textbf{Framework:} We evaluate every model as a tool-using agent in a single ReAct~\cite{react} loop implemented with LangChain. Each of the 128 GIS APIs (Section~\ref{sec:dataset}) is registered as a typed tool, so every model sees the same interface. The agent alternates between reasoning and acting until it declares completion or exhausts its step budget, and every tool call, argument, return value, and error is logged.

\noindent\textbf{Metrics:} For model output errors, we use strict \textbf{Task Success Rate (TSR)}, the fraction of tasks whose outputs match the ground truth within a small tolerance threshold, and \textbf{Quantitative Closeness Score (QCS)}, which aggregates closeness across both non-spatial and spatial attributes, with a missing output scoring $0$. Non-spatial attributes include qualitative (weighted $F_1$) and quantitative ($1-\mathrm{sMAPE}$, symmetric mean absolute percentage error). Rows are matched by the output contract's identifier columns rather than by position, and attribute scores are computed over matched rows only. Spatial attributes include raster grids and multi-point, multi-polyline, and multi-polygon geometries, which is measured by instance $F_1$ within an intersection-over-union (IoU) threshold for areal outputs or a distance threshold for point and line outputs. We additionally report \textbf{trajectory closeness}, which measures the similarity of key API calls at a coarse functional category level. We also record trajectory length and tool-call error rate. Appendix~F gives full definitions.

\noindent\textbf{Infrastructure:} Agent execution and all geoprocessing run in a reproducible container on one node with 32 CPU cores and 512\,GB of memory. No GPUs are used: all six models are accessed as hosted API services. Appendix~A reports the software environment, decoding parameters, and model routing.

\noindent\textbf{Models:} We evaluate six agents---three closed-source models (Gemini-3.1-Pro, Claude-4.6-Opus, GPT-5.4) and three open-source models (DeepSeek-V4-Pro, Qwen3.6-27B, GPT-OSS-120B)---on all 349 tasks. All six run under the same harness, prompts, and 50-call budget with greedy decoding, so measured differences reflect the models rather than their scaffolding.

\subsection{Overall Output Accuracy Analysis}

Table~\ref{tab:overall_results} summarizes overall model performance and output accuracy across the benchmark. No agent solves a third of the tasks: Gemini-3.1-Pro leads at 0.327 strict TSR against a six-model mean of 0.238, the three closed-source models form a clear tier, and the open-source models reach roughly half their success rate. Quantitative closeness preserves that ranking but is two to three times higher throughout (0.380--0.663), so a typical failed run recovers much of the intended result rather than producing nothing.

Difficulty is driven by workflow type rather than model-specific strengths: the family ranking is consistent across all six models. Spatial Overlay \& Suitability Analysis is most tractable and Spatial Pattern Analysis hardest under both metrics, its low closeness score indicating genuinely wrong outputs rather than near misses. Terrain Modeling inverts this---the second-lowest TSR but the highest quantitative closeness of any family---because long raster pipelines run to completion and land just outside tolerance.

Numeric closeness and categorical agreement are uniformly high and separate the models weakly (Tables~\ref{tab:nonspatial_errors} and~\ref{tab:spatial_errors}); the discriminative signals are row coverage and geometry. Spatial quality is comparatively strong---entity $F_1$ reaches 0.76--0.89---so agents recover shapes far more reliably than they recover complete, correctly valued row sets. Strict failures are therefore predominantly attribute- and coverage-level rather than geometric.

\begin{table}[t]
\centering
\caption{Non-spatial output accuracy.}
\label{tab:nonspatial_errors}
\small
\setlength{\tabcolsep}{3pt}
\begin{tabular}{lccc}
\toprule
& \textbf{Numerical} & \textbf{Categorical} & \textbf{Row} \\
\textbf{Model} & $\mathbf{1-}$\textbf{sMAPE} & \textbf{Weighted F1} & \textbf{Coverage} \\
\midrule
Gemini-3.1-Pro   & \textbf{0.903} & 0.980 & 0.871 \\
Claude-4.6-Opus  & 0.859 & 0.983 & 0.866 \\
DeepSeek-V4-Pro  & 0.888 & \textbf{0.984} & \textbf{0.890} \\
GPT-5.4          & 0.878 & 0.945 & 0.868 \\
Qwen3.6-27B      & 0.885 & 0.970 & 0.844 \\
GPT-OSS-120B     & 0.880 & 0.963 & 0.833 \\
\bottomrule
\end{tabular}
\end{table}
\begin{table}[t]
\centering
\caption{Spatial output accuracy by entity-level matching.}
\label{tab:spatial_errors}
\small
\setlength{\tabcolsep}{3pt}
\begin{tabular}{lccc ccc}
\toprule
& \multicolumn{3}{c}{\textbf{All spatial entities}} & \multicolumn{3}{c}{\textbf{F1 by geometry type}} \\
\cmidrule(lr){2-4} \cmidrule(lr){5-7}
\textbf{Model} & \textbf{Prec.} & \textbf{Rec.} & \textbf{F1} & \textbf{Point} & \textbf{Line} & \textbf{Polygon} \\
\midrule
Gemini-3.1-Pro   & 0.886 & 0.889 & 0.876 & 0.886 & \textbf{0.995} & 0.868 \\
Claude-4.6-Opus  & 0.899 & \textbf{0.900} & 0.882 & \textbf{0.928} & 0.857 & 0.873 \\
DeepSeek-V4-Pro  & \textbf{0.901} & 0.899 & \textbf{0.888} & 0.925 & 0.662 & \textbf{0.895} \\
GPT-5.4          & 0.821 & 0.797 & 0.797 & 0.723 & 0.800 & 0.806 \\
Qwen3.6-27B      & 0.883 & 0.847 & 0.855 & 0.835 & \textbf{1.000} & 0.843 \\
GPT-OSS-120B     & 0.762 & 0.786 & 0.771 & 0.781 & 0.750 & 0.762 \\
\bottomrule
\end{tabular}
\end{table}

\subsection{Agentic Trajectory Evaluation}

\begin{figure}[t]
\centering
\includegraphics[width=0.72\columnwidth]{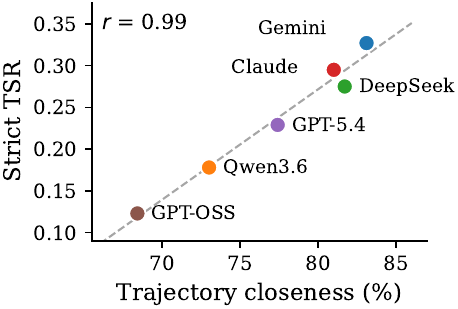}
\caption{Strict TSR against trajectory closeness. Agents cover most required operations but convert few of them into exactly correct outputs.}
\label{fig:tsr_traj}
\end{figure}

Figure~\ref{fig:tsr_traj} plots trajectory closeness against strict TSR. The two are consistent with each other: models that perform more of the required operations generally solve more tasks, and the same three models lead on both measures ($r=0.99$). They nevertheless sit on very different scales, with agents covering 68--83\% of the required operations while solving 12--33\% of tasks.

We report trajectory closeness as a diagnostic rather than as a headline metric, and retain output error analysis as our primary measure: scoring a produced output against an exact ground truth is what distinguishes this benchmark from prior work that can only compare trajectories. A formal trajectory metric would need more than a correct reference trajectory, since a task admits many equally valid trajectories that differ in the order of independent steps and in the choice among interchangeable operations. Defining equivalence over that space is left to future work.

Execution statistics of model trajectories are given in Table~\ref{tab:exec_stats}. Average trajectory length (number of tool calls) ranges from 13.6 to 28.5. Qwen3.6027B has bigger tool call error rates, likely due to a smaller model size. The rankings of tool call error rate is also consistent with output errors.

\begin{table}[b]
\centering
\caption{Execution statistics: mean tool calls per task and the fraction of tool calls returning an error.}
\label{tab:exec_stats}
\small
\begin{tabular}{lcc}
\toprule
\textbf{Model} & \textbf{Mean tool calls} & \textbf{Tool call error rate} \\
\midrule
Gemini-3.1-Pro   & 25.5 & \textbf{2.9\%} \\
Claude-4.6-Opus  & 15.8 & 3.1\% \\
DeepSeek-V4-Pro  & 28.5 & 4.3\% \\
GPT-5.4          & \textbf{13.6} & 4.0\% \\
Qwen3.6-27B      & 22.4 & 12.0\% \\
GPT-OSS-120B     & 22.1 & 9.5\% \\
\bottomrule
\end{tabular}
\end{table}

\begin{figure}[t]
\centering
\includegraphics[width=0.82\columnwidth]{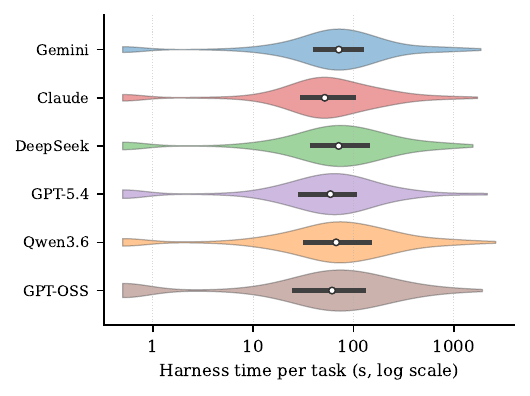}
\caption{Per-task harness time by model, log scale. Bars mark the interquartile range, dots the median.}
\label{fig:harness_time}
\end{figure}

\subsection{Computational Cost of API Calls}

Figure~\ref{fig:harness_time} shows how long each task takes to execute inside the harness. We exclude the LLM API time costs due to inconsistency of cloud API providers. The results show that task API time cost is heavy-tailed: it spans four orders of magnitude, from sub-second attribute edits to a 2{,}633\,s spatial join over 939{,}821 address points, so the slowest 5\% of runs consume 38\% of all execution time and the mean sits at roughly twice the median. Models differ only modestly in the execution they trigger, from a median of 52\,s per task for Claude-4.6-Opus to 72\,s for Gemini-3.1-Pro, a spread of $1.38\times$. We also measure the {\bf token consumption} of different models, from 275k per task for GPT-5.4 to 829k for DeepSeek-V4-Pro. Appendix~G and~H give the per-family breakdowns.

\subsection{Failure Analysis}

Each incorrect run with a trajectory carries a written diagnostic note from scoring. We labelled those 1{,}381 notes against a fixed set of mechanisms, allowing more than one label per run; Table~\ref{tab:failure_causes} reports how often each appears.

Failures are concentrated in the plan rather than in the individual call. A required operation is missing from 28.3\% of runs, and another 18.4\% perform the right operations in the wrong order, most often resampling a raster after computing a slope from it rather than before. Wrong tool substitution (13.4\%) is similar in kind: the agent picks an operation that sounds right but computes something else, typically a plain spatial join where one that aggregates was needed, and a further 14.4\% repeat or undo steps without converging. Errors inside a single call are rarer, at 7.8\%, while 9.9\% of runs compute much of the answer but never write the required file.

Three rows describe the limits of the analysis rather than agent behaviour. \emph{Complete but unexplained} (14.1\%) covers runs that perform every required operation in a valid order yet still produce a wrong output, with no cause visible in the sequence of tool names; \emph{shared task-level ceiling} (13.8\%) covers tasks every model fails alike, which points at the task rather than the agent; and 16.8\% of notes could not be assigned to any mechanism.

\noindent\textbf{Performance by caveat.} The data-integrity caveats are the most damaging: pooled TSR falls to 0.152 on tasks tagged with geometry and topology errors and 0.220 under CRS misalignment, against 0.241 for NoData handling and 0.269--0.277 for the language-level caveats (boundary ambiguity and unit mismatch). Underspecified predicates produce syntactically valid but semantically wrong outputs rather than failed calls, so they cost accuracy without raising the error rate.

\begin{table}[t]
\centering
\caption{Failure mechanisms across 1{,}381 wrong-but-attempted runs, pooled over all six models.}
\label{tab:failure_causes}
\small
\setlength{\tabcolsep}{4pt}
\begin{tabular}{L{5.4cm}rr}
\toprule
\textbf{Failure mechanism} & \textbf{Runs} & \textbf{\%} \\
\midrule
Missing required operation      & 391 & 28.3 \\
Operation order violation       & 254 & 18.4 \\
Redundant rework                & 199 & 14.4 \\
Complete but unexplained        & 195 & 14.1 \\
Shared task-level ceiling       & 190 & 13.8 \\
Wrong tool substitution         & 185 & 13.4 \\
Output never delivered          & 137 &  9.9 \\
Parameter or formula error      & 108 &  7.8 \\
Unrecovered tool failure        &  47 &  3.4 \\
Native identifier violation     &  38 &  2.8 \\
CRS / reprojection gap          &  38 &  2.8 \\
Premature or unsorted delivery  &  26 &  1.9 \\
\midrule
Not specified                   & 232 & 16.8 \\
\bottomrule
\end{tabular}
\end{table}

\subsection{Case Studies}

We conducted case studies for representative success and failure examples for each task family, with map visualizations of agent and reference outputs. Due to space limit, they are provided in Appendix~I.

\section{Conclusion and Future Work}
\label{sec:conclusion}

We introduced {GISAgentBench}, a benchmark for evaluating LLM agents on the multi-step spatial analysis that GIS professionals perform in practice. Its 349 tasks come from questions working analysts asked on a practitioner forum, recast onto new geography so that no published answer exists for any of them, and each ships with real spatial data, an executable reference solution, a ground truth output file, and expert annotations marking the pitfalls that cost practitioners hours of debugging. Agents work through a fixed harness of 128 GIS APIs and are scored deterministically against the reference within task-declared tolerances, with no model in the loop. Across six LLM agents, the strongest solves 32.7\% of tasks fully, and failures are dominated by omitted and mis-ordered operations rather than malformed calls, evidence that capability is limited by workflow planning rather than by access to GIS functionality.

Future work will extend the harness beyond open-source libraries to commercial GIS libraries such as ArcGIS Toolbox, which reach operations our present API set does not. We will broaden geographic coverage beyond the predominantly U.S. areas of interest used here. We will also evaluate new frontier models as they are released, so that the benchmark tracks the capability frontier rather than a single snapshot of it. Evaluating open-ended and purely cartographic products, which deterministic ground truth output matching cannot score, remains open.

\bibliography{ref}

\clearpage
\onecolumn
\twocolumn
\setcounter{section}{0}
\setcounter{table}{0}
\setcounter{figure}{0}
\renewcommand{\thesection}{\Alph{section}}
\renewcommand{\thetable}{\Alph{section}\arabic{table}}
\renewcommand{\thefigure}{\Alph{section}\arabic{figure}}

\section*{Appendices}
\noindent
The appendices below record the implementation and experimental details needed
to reproduce our results, together with extended tables and analyses for which
there was no room in the main text.

\section{Reproducibility Details}
\label{app:repro}

\subsection{Model Access and Decoding}

All six evaluated models are accessed as hosted API services; none is served on local hardware, so no GPU resources are used for inference. Gemini-3.1-Pro, Claude-4.6-Opus, GPT-5.4, and GPT-OSS-120B are reached through an institutional API gateway that proxies vendor endpoints, while DeepSeek-V4-Pro and Qwen3.6-27B are reached through OpenRouter. All runs reported in the paper were executed in July 2026.

Table~\ref{tab:hyperparams} lists every parameter used. Decoding is greedy: temperature is set explicitly to $0$ for all six models, and no other sampling parameter (top-$p$, top-$k$, or repetition penalty) is overridden, so provider defaults apply. We do not set a random seed, because the hosted endpoints do not expose seed control; temperature-$0$ decoding is the only determinism mechanism available to us. Exact bitwise replication is therefore not guaranteed for hosted models, and providers may additionally update model versions behind a fixed endpoint name. We perform one run per task per model, so all reported numbers are single-run values rather than averages over repetitions.

\begin{table}[t]
\centering
\small
\setlength{\tabcolsep}{4pt}
\caption{Final parameter values, applied identically to all six models. No hyperparameter search was performed.}
\label{tab:hyperparams}
\begin{tabular}{lc}
\toprule
\textbf{Parameter} & \textbf{Value} \\
\midrule
Temperature                       & 0 (greedy) \\
Top-$p$, top-$k$, penalties       & provider default \\
Random seed                       & not exposed by providers \\
Max tool calls per task           & 50 \\
Max tool-result characters        & 4{,}000 \\
Runs per task per model           & 1 \\
Agent loop                        & ReAct~\cite{react} \\
Toolset size                      & 128 primitives \\
\bottomrule
\end{tabular}
\end{table}

\subsection{Software and Hardware Environment}

Agent execution and all geoprocessing run inside a reproducible container on a single compute node with 32 CPU cores and 512\,GB of memory; no GPUs are used. The software environment is Ubuntu 25.10 with Python 3.11.15, QGIS 4.0.2, GDAL/OGR 3.10.3, PROJ 9.6.0, GEOS 3.13.1, GeoPandas 1.1.3, Rasterio 1.4.4, Shapely 2.1.2, and PyProj 3.7.2. Because every benchmark task is scored by deterministic output matching against a stored ground truth file, evaluation is fully reproducible given a fixed set of agent trajectories, independent of the model endpoints used to generate them.

\section{Task Set Construction Details}
\label{app:construction}

This appendix records the concrete filtering criteria and prompts behind the pipeline summarized in Section~2.1 of the main paper.

\subsection{Source and Crawl}

Threads are collected from GIS Stack Exchange through the Stack Exchange API, retaining the question title, body, tags, and answer bodies. The crawler also implements a scraper for a second practitioner forum, which we did not use for this release; every task in GISAgentBench originates from GIS Stack Exchange.

\subsection{Deterministic Pre-Filter}

The first stage uses no language model. Each thread is scored against seven positive term dictionaries and eight blacklist families, all compiled as word-boundary regular expressions. The dictionaries are reproduced in full below, exactly as they appear in the pipeline source.

\paragraph{Positive dictionaries.}
\begin{description}\small\setlength{\itemsep}{2pt}
\item[\textbf{GIS core terminology} (53).] \emph{arcgis}, \emph{arcgis pro}, \emph{arcmap}, \emph{arcpy}, \emph{arcpy.analysis}, \emph{arcpy.conversion}, \emph{arcpy.da}, \emph{arcpy.management}, \emph{arcpy.mp}, \emph{arcpy.na}, \emph{arcpy.sa}, \emph{contour}, \emph{coordinate system}, \emph{crs}, \emph{datum}, \emph{datum transformation}, \emph{dem}, \emph{epsg}, \emph{feature class}, \emph{feature layer}, \emph{featureclass}, \emph{file geodatabase}, \emph{geodatabase}, \emph{geographic coordinate system}, \emph{geometry}, \emph{geoprocess}, \emph{geoprocessing}, \emph{geotiff}, \emph{gis}, \emph{hydrology}, \emph{image analyst}, \emph{lidar}, \emph{map algebra}, \emph{nad83}, \emph{network analyst}, \emph{parcel}, \emph{point}, \emph{polygon}, \emph{polyline}, \emph{projected coordinate system}, \emph{projection}, \emph{raster}, \emph{shapefile}, \emph{spatial analyst}, \emph{spatial reference}, \emph{tin}, \emph{topology}, \emph{utm}, \emph{vector}, \emph{watershed}, \emph{web mercator}, \emph{wgs84}, \emph{wkid}.
\item[\textbf{Workflow verbs} (54).] \emph{add field}, \emph{append}, \emph{aspect}, \emph{buffer}, \emph{calculate field}, \emph{clip}, \emph{contour}, \emph{copy features}, \emph{cost distance}, \emph{cost path}, \emph{define projection}, \emph{dissolve}, \emph{erase}, \emph{euclidean distance}, \emph{extract by mask}, \emph{extract values to points}, \emph{feature class to feature class}, \emph{flow accumulation}, \emph{flow direction}, \emph{focal statistics}, \emph{generate near table}, \emph{idw}, \emph{intersect}, \emph{join field}, \emph{kernel density}, \emph{kriging}, \emph{least cost path}, \emph{make feature layer}, \emph{make xy event layer}, \emph{merge}, \emph{mosaic to new raster}, \emph{near}, \emph{pairwise buffer}, \emph{pairwise clip}, \emph{pairwise erase}, \emph{pairwise intersect}, \emph{point density}, \emph{project raster}, \emph{raster calculator}, \emph{reclassify}, \emph{reproject}, \emph{sample}, \emph{select layer by attribute}, \emph{select layer by location}, \emph{slope}, \emph{snap raster}, \emph{spatial join}, \emph{stream order}, \emph{summarize within}, \emph{table to table}, \emph{tabulate intersection}, \emph{union}, \emph{viewshed}, \emph{zonal statistics}.
\item[\textbf{Network analysis} (19).] \emph{add locations}, \emph{closest facility}, \emph{impedance}, \emph{location allocation}, \emph{make closest facility analysis layer}, \emph{make closest facility layer}, \emph{make od cost matrix analysis layer}, \emph{make route analysis layer}, \emph{make service area layer}, \emph{network dataset}, \emph{od cost matrix}, \emph{origin destination cost matrix}, \emph{route}, \emph{routing}, \emph{service area}, \emph{shortest path}, \emph{solve tool}, \emph{travel mode}, \emph{turn restriction}.
\item[\textbf{GIS API terms} (21).] \emph{addfield}, \emph{arcpy.describe}, \emph{arcpy.exists}, \emph{arcpy.getcount}, \emph{arcpy.listfeatureclasses}, \emph{arcpy.listfields}, \emph{arcpy.management.getcount}, \emph{calculatefield}, \emph{copyfeatures}, \emph{getcount}, \emph{insertcursor}, \emph{listdatasets}, \emph{listfeatureclasses}, \emph{listfields}, \emph{listtables}, \emph{makefeaturelayer}, \emph{searchcursor}, \emph{selectlayerbyattribute}, \emph{selectlayerbylocation}, \emph{spatialreference}, \emph{updatecursor}.
\item[\textbf{Data context} (27).] \emph{.csv}, \emph{.gdb}, \emph{.geojson}, \emph{.gpkg}, \emph{.img}, \emph{.shp}, \emph{.tif}, \emph{.tiff}, \emph{address}, \emph{addresses}, \emph{attribute table}, \emph{building footprint}, \emph{field name}, \emph{field names}, \emph{latitude}, \emph{line layer}, \emph{longitude}, \emph{objectid}, \emph{point layer}, \emph{polygon layer}, \emph{road network}, \emph{shape@}, \emph{shape@xy}, \emph{sql}, \emph{street centerline}, \emph{timestamp}, \emph{where clause}.
\item[\textbf{Task-signal verbs} (28).] \emph{aggregate}, \emph{buffer}, \emph{calculate}, \emph{classify}, \emph{clip}, \emph{compute}, \emph{convert}, \emph{create}, \emph{derive}, \emph{estimate}, \emph{extract}, \emph{find}, \emph{geocode}, \emph{identify}, \emph{intersect}, \emph{join}, \emph{overlay}, \emph{produce}, \emph{project}, \emph{repair}, \emph{reverse geocode}, \emph{route}, \emph{sample}, \emph{segment}, \emph{summarize}, \emph{tabulate}, \emph{transform}, \emph{validate}.
\item[\textbf{Result-signal terms} (18).] \emph{classification}, \emph{count}, \emph{csv}, \emph{density}, \emph{feature class}, \emph{feature layer}, \emph{hot spot}, \emph{layer}, \emph{mean}, \emph{output}, \emph{point}, \emph{polygon}, \emph{polyline}, \emph{raster}, \emph{result}, \emph{route}, \emph{summary}, \emph{table}.
\end{description}

\paragraph{Blacklist families.}
\begin{description}\small\setlength{\itemsep}{2pt}
\item[\textbf{Not a task} (21).] \emph{account}, \emph{announcement}, \emph{career}, \emph{community guidelines}, \emph{enhancement}, \emph{feature request}, \emph{hiring}, \emph{idea}, \emph{internship}, \emph{job posting}, \emph{license}, \emph{licensing}, \emph{login}, \emph{moderator}, \emph{pricing}, \emph{release notes}, \emph{roadmap}, \emph{server outage}, \emph{service issue}, \emph{status page}, \emph{subscription}.
\item[\textbf{Interface only} (19).] \emph{display}, \emph{export format}, \emph{hatch fill}, \emph{html margin}, \emph{icon}, \emph{label}, \emph{labeling}, \emph{layout}, \emph{legend}, \emph{panel}, \emph{pop up}, \emph{popup}, \emph{print layout}, \emph{render}, \emph{ribbon}, \emph{style only}, \emph{styling}, \emph{symbology}, \emph{toolbar}.
\item[\textbf{Environment and installation} (14).] \emph{cannot load}, \emph{conda}, \emph{corrupt package}, \emph{dependency}, \emph{environment}, \emph{failed to load}, \emph{install}, \emph{installation}, \emph{package}, \emph{packages}, \emph{pip}, \emph{plugin}, \emph{upgrade}, \emph{version}.
\item[\textbf{Online services} (12).] \emph{access denied}, \emph{arcgis online}, \emph{geoserver}, \emph{not running}, \emph{permission}, \emph{portal}, \emph{publish}, \emph{publishing}, \emph{refresh}, \emph{service unavailable}, \emph{submitted but not running}, \emph{sync}.
\item[\textbf{Software recommendation} (10).] \emph{basemap provider}, \emph{best software}, \emph{best tool}, \emph{imagery provider}, \emph{recommend a}, \emph{recommend software}, \emph{recommendation}, \emph{tile provider}, \emph{what software}, \emph{which software}.
\item[\textbf{General questions} (5).] \emph{general question}, \emph{is there a way}, \emph{what is better}, \emph{what is the difference}, \emph{why does}.
\item[\textbf{Weak help requests} (10).] \emph{any ideas}, \emph{can anyone help}, \emph{can someone help}, \emph{does anyone know}, \emph{i am stuck}, \emph{i need help}, \emph{not sure why}, \emph{please help}, \emph{thanks in advance}, \emph{what am i doing wrong}.
\item[\textbf{Code errors} (10).] \emph{attributeerror}, \emph{exception:}, \emph{indexerror}, \emph{keyerror}, \emph{runtimeerror}, \emph{stack trace}, \emph{syntaxerror}, \emph{traceback}, \emph{typeerror}, \emph{valueerror}.
\end{description}

A thread's keyword score sums its dictionary hits and the number of task-category families it matches, where a family matches when the thread contains at least two of its subject terms, or at least one subject and one action term. Threads are rejected when the problem statement is shorter than 420 characters, when the keyword score falls below 4, when no task family matches and the thread lacks both a workflow signal and two or more data-context terms, when no task signal or result signal is present, or when fewer than two data-context terms appear. Near-duplicates are removed by a SHA-256 signature over the lowercased title and the first 1{,}200 characters of the statement.

\subsection{LLM Screening}

Two GPT-5-mini screening passes follow. The first is a cheap triage that emits a keep/reject decision against these criteria, quoted verbatim from the pipeline:

\begin{quote}\small\itshape
Reject a thread if any line below fits it. (1)~It is not a GIS work problem. (2)~It is mainly about accounts, licenses, access, setup, installs, packages, plugins, or versions. (3)~It is mainly about display, colors, labels, legends, layouts, popups, forms, styling, settings, or map appearance. (4)~It is mainly about code syntax, SQL syntax, API syntax, script bugs, or tool settings instead of the GIS result. (5)~It is mainly about file import/export, file format quirks, metadata, storage, or docs. (6)~It asks for general advice or explanation, not a concrete GIS output. (7)~It is only a tiny one-step conversion, lookup, report, extract, or cleanup. (8)~It is mainly about speed, hardware, compression, parallel work, or rendering. (9)~The GIS goal is not clear.
\end{quote}

The second pass is stricter and decides whether a surviving thread can become a benchmark seed:

\begin{quote}\small\itshape
Keep only threads that can become strong GIS benchmark seeds. Keep when all lines below are true: (1)~The post has a real GIS task, not just a bug, UI issue, or software question. (2)~The task has a clear spatial goal and a clear final result. (3)~The task can be written so the same inputs always lead to the same output. (4)~The task can be moved to local GIS files without changing the real problem. (5)~The work is more than one tiny tool call. (6)~The post or answer gives enough detail to write a GIS solution later. Reject if a later task writer would need to invent a different goal, guess missing facts, or turn the post into a new problem.
\end{quote}

This pass returns structured output rather than a bare label: booleans recording whether the thread states a complete problem, whether its original meaning can be preserved under recasting, and whether a short reference trajectory is plausible; together with integer ratings on complexity, usability, reproducibility, ambiguity, and reference clarity.

\subsection{Task Drafting}

Retained threads are cast into the benchmark task template: a task identifier, a practitioner-style prompt with title, objective, inputs, required outputs and constraints, the input roles bound to concrete files in a selected area of interest, task parameters, and an output contract naming the file kind, required columns, spatial reference, and tolerances. The drafting stage preserves the original analytical objective and is explicitly barred from inventing a different question; drafts that would require doing so are rejected rather than repaired. Every draft is then reviewed by a second model pass before entering reference generation.

The drafting instruction is quoted verbatim:

\begin{quote}\small\itshape
Create one local GIS benchmark task from one GIS forum thread.
Return status=ready only when the task is clear, local, checkable, and still the same real problem as the source post.
Reject if the task needs missing data, guessed input columns, made-up raster class meanings, made-up selected features, fake row pairs, fake ObjectID matches, unsupported network work, manual choices, or a different GIS problem.
Return exactly one JSON object matching the schema.
\end{quote}

A draft that fails schema or style validation is passed to a repair prompt, which may correct fields but is held to the same prohibition on inventing a different question.

\section{Complete GIS API Harness}
\label{app:apis}

Table~\ref{tab:api_full} lists all 128 APIs available to the agent, grouped by function. Descriptions are the first line of each API's docstring, which is verbatim the description the agent receives at inference time, so the table reflects the harness exactly as the models see it. Of these, 111 appear in at least one released reference trajectory; the remainder are available to agents but not required by any task.

APIs are backed by QGIS through the \texttt{qgis\_process} command-line interface, including its GDAL-backed algorithms, together with GeoPandas, Rasterio, Shapely, pandas/NumPy, and SciPy. Roughly two thirds dispatch to a QGIS processing algorithm; the remainder are implemented directly against the Python geospatial stack, and a small number compose other APIs in the harness.

\onecolumn
{\footnotesize
\setlength{\LTleft}{0pt}\setlength{\LTright}{0pt}\setlength{\LTcapwidth}{0.95\textwidth}
\captionof{table}{The complete 128-API harness. For each API the parameter list is given beneath its name, with optional parameters in brackets. Descriptions are the first line of each API's docstring, which is verbatim what the agent receives as the tool description.}\label{tab:api_full}
\begin{longtable}{@{}>{\raggedright\arraybackslash}p{0.38\textwidth}>{\raggedright\arraybackslash}p{0.57\textwidth}@{}}
\toprule \textbf{API and parameters} & \textbf{Description} \\ \midrule \endfirsthead
\multicolumn{2}{@{}l}{\itshape Table \thetable\ continued from previous page}\\
\toprule \textbf{API and parameters} & \textbf{Description} \\ \midrule \endhead
\bottomrule \endlastfoot
\multicolumn{2}{@{}l}{\textbf{Inspection \& metadata (read-only)} (6)}\\[0.2em]
\texttt{describe\_\allowbreak layer}\newline {\scriptsize\itshape INPUT} & Return schema and spatial metadata for a vector layer or CSV. \\
\texttt{describe\_\allowbreak raster}\newline {\scriptsize\itshape INPUT} & Return spatial metadata for a raster. \\
\texttt{get\_\allowbreak feature\_\allowbreak count}\newline {\scriptsize\itshape INPUT} & Return the total number of features (rows) in a vector layer or CSV. \\
\texttt{get\_\allowbreak field\_\allowbreak unique\_\allowbreak values}\newline {\scriptsize\itshape INPUT, FIELD, [MAX\_VALUES]} & Return the sorted unique non-null values of FIELD, capped at MAX\_VALUES. \\
\texttt{get\_\allowbreak feature\_\allowbreak id\_\allowbreak field}\newline {\scriptsize\itshape INPUT} & Return the QGIS expression string '\$id'. Pass the returned value as FORMULA in calculate\_attribute\_field to materialize a 1-based sequential integer feature identifier into a named\textbackslash dots \\
\texttt{get\_\allowbreak raster\_\allowbreak property}\newline {\scriptsize\itshape INPUT, PROPERTY\_NAME} & Return one scalar property of a raster dataset. \\
\addlinespace
\multicolumn{2}{@{}l}{\textbf{CRS \& reprojection} (4)}\\[0.2em]
\texttt{warp\_\allowbreak reproject\_\allowbreak raster}\newline {\scriptsize\itshape INPUT, TARGET\_CRS, OUTPUT, [RESAMPLING], [SOURCE\_CRS], [TARGET\_RESOLUTION], [TARGET\_EXTENT], [TARGET\_EXTENT\_CRS], [NODATA], [DATA\_TYPE], [MULTITHREADING], [EXTRA]} & Reproject or resample a raster to TARGET\_CRS using GDAL warp. \\
\texttt{assign\_\allowbreak projection}\newline {\scriptsize\itshape INPUT, CRS, OUTPUT} & Assign CRS metadata to INPUT without reprojecting coordinates. \\
\texttt{reproject\_\allowbreak features}\newline {\scriptsize\itshape INPUT, TARGET\_CRS, OUTPUT, [CONVERT\_CURVED\_GEOMETRIES], [OPERATION]} & Transform vector layer coordinates to TARGET\_CRS (e.g. 'EPSG:28992'). \\
\texttt{align\_\allowbreak raster\_\allowbreak to\_\allowbreak template}\newline {\scriptsize\itshape INPUT, TEMPLATE, OUTPUT, [RESAMPLING], [DATA\_TYPE], [NODATA], [MULTITHREADING]} & Resample INPUT to exactly match TEMPLATE's CRS, origin, cell size, and pixel dimensions. \\
\addlinespace
\multicolumn{2}{@{}l}{\textbf{Vector overlay} (8)}\\[0.2em]
\texttt{intersection\_\allowbreak overlay}\newline {\scriptsize\itshape INPUT, OVERLAY, OUTPUT, [INPUT\_FIELDS], [OVERLAY\_FIELDS], [OVERLAY\_FIELDS\_PREFIX], [GRID\_SIZE]} & Compute the geometric intersection of INPUT and OVERLAY, producing output features with the intersecting geometry and attributes from both layers. \\
\texttt{clip\_\allowbreak features}\newline {\scriptsize\itshape INPUT, OVERLAY, OUTPUT} & Clip INPUT geometries to the shape of the OVERLAY layer, discarding parts that fall outside. INPUT attributes are preserved. \\
\texttt{difference\_\allowbreak overlay}\newline {\scriptsize\itshape INPUT, OVERLAY, OUTPUT, [GRID\_SIZE]} & Erase OVERLAY geometry from INPUT features, keeping only the parts of INPUT that do not overlap OVERLAY. INPUT attributes are preserved. \\
\texttt{line\_\allowbreak intersections}\newline {\scriptsize\itshape INPUT, INTERSECT, OUTPUT, [INPUT\_FIELDS], [INTERSECT\_FIELDS], [INTERSECT\_FIELDS\_PREFIX]} & Find all crossing points between INPUT and INTERSECT line layers, outputting point features with attributes from both. \\
\texttt{union\_\allowbreak overlay}\newline {\scriptsize\itshape INPUT, OUTPUT, [OVERLAY], [OVERLAY\_FIELDS\_PREFIX], [GRID\_SIZE]} & Compute the geometric union of INPUT and OVERLAY, producing output features that cover all areas of both layers with attributes from both. When OVERLAY is omitted, dissolves INPUT into a\textbackslash dots \\
\texttt{multi\_\allowbreak intersection\_\allowbreak overlay}\newline {\scriptsize\itshape INPUT, OVERLAYS, OUTPUT, [OVERLAY\_FIELDS\_PREFIX]} & Compute the geometric intersection of INPUT with every layer in OVERLAYS sequentially, retaining attributes from all layers. \\
\texttt{multi\_\allowbreak union\_\allowbreak overlay}\newline {\scriptsize\itshape INPUT, OVERLAYS, OUTPUT, [OVERLAY\_FIELDS\_PREFIX]} & Compute the geometric union of INPUT with every layer in OVERLAYS, retaining all features and attributes from all layers. \\
\texttt{update\_\allowbreak features}\newline {\scriptsize\itshape INPUT, UPDATE, OUTPUT, [KEEP\_BORDERS], [CLUSTER\_TOLERANCE]} & Replace INPUT areas overlapping UPDATE features with the UPDATE features, then merge the remainder with UPDATE into a single output layer. \\
\addlinespace
\multicolumn{2}{@{}l}{\textbf{Buffers \& proximity} (10)}\\[0.2em]
\texttt{buffer\_\allowbreak features}\newline {\scriptsize\itshape INPUT, DISTANCE, OUTPUT, [SEGMENTS], [END\_CAP\_STYLE], [JOIN\_STYLE], [MITER\_LIMIT], [DISSOLVE]} & Create buffer polygons at radius DISTANCE around each input feature. \\
\texttt{multi\_\allowbreak ring\_\allowbreak buffer}\newline {\scriptsize\itshape INPUT, DISTANCES, FIELD\_NAME, OUTPUT, [OUTSIDE\_ONLY], [DISSOLVE]} & Create concentric buffer polygons at each distance in DISTANCES from input features, storing the ring distance in FIELD\_NAME. \\
\texttt{nearest\_\allowbreak features\_\allowbreak table}\newline {\scriptsize\itshape INPUT, NEAR, OUTPUT, [SEARCH\_RADIUS], [LOCATION], [ANGLE], [NEIGHBOR\_LIMIT], [DISTANCE\_METHOD], [DROP\_GEOMETRY], [INPUT\_ID\_FIELD], [NEAR\_ID\_FIELD]} & Computes nearest-neighbor distances between INPUT and NEAR features and writes a table to OUTPUT. Output columns: IN\_FID (INPUT identifier), NEAR\_FID (NEAR identifier), NEAR\_DIST\textbackslash dots \\
\texttt{nearest\_\allowbreak feature\_\allowbreak distance\_\allowbreak layer}\newline {\scriptsize\itshape INPUT, NEAR, OUTPUT, [INPUT\_ID\_FIELD], [NEAR\_ID\_FIELD], [OUTPUT\_INPUT\_ID\_FIELD], [OUTPUT\_NEAR\_ID\_FIELD], [OUTPUT\_DISTANCE\_FIELD], [ROUND\_DECIMALS], [DISTANCE\_MULTIPLIER], [KEEP\_FIELDS]} & For each INPUT feature, finds the nearest NEAR feature and writes OUTPUT: a spatial layer with INPUT geometry and three appended fields: OUTPUT\_INPUT\_ID\_FIELD (INPUT feature identifier),\textbackslash dots \\
\texttt{euclidean\_\allowbreak distance\_\allowbreak raster}\newline {\scriptsize\itshape SOURCE, OUTPUT, [MAXIMUM\_DISTANCE], [CELL\_SIZE]} & Create a raster where each cell value is the planar Euclidean distance to the nearest non-zero cell in SOURCE. SOURCE must be a raster with non-zero cells marking the source locations. \\
\texttt{buffer\_\allowbreak polygon\_\allowbreak area\_\allowbreak perimeter\_\allowbreak table}\newline {\scriptsize\itshape POINTS, POLYGONS, OUTPUT, [POINT\_ID\_FIELD], [BUFFER\_RADIUS], [OUTPUT\_AREA\_FIELD], [OUTPUT\_PERIMETER\_FIELD], [OUTPUT\_PERCENT\_FIELD], [ROUND\_DECIMALS]} & For each point, compute the intersection area, boundary length, and area percentage of POLYGONS within a BUFFER\_RADIUS circle. Output columns: POINT\_ID\_FIELD, OUTPUT\_AREA\_FIELD (total\textbackslash dots \\
\texttt{polygon\_\allowbreak max\_\allowbreak interior\_\allowbreak radius\_\allowbreak table}\newline {\scriptsize\itshape INPUT, OUTPUT, [ID\_FIELD], [OUTPUT\_ID\_FIELD], [OUTPUT\_RADIUS\_FIELD], [TOLERANCE], [ROUND\_DECIMALS]} & Compute the pole of inaccessibility radius (maximum inscribed circle radius) for each polygon and write a table. Output columns: OUTPUT\_ID\_FIELD (polygon identifier from ID\_FIELD),\textbackslash dots \\
\texttt{clip\_\allowbreak point\_\allowbreak buffers\_\allowbreak to\_\allowbreak containing\_\allowbreak polygons}\newline {\scriptsize\itshape POINTS, POLYGONS, OUTPUT, [POINT\_ID\_FIELD], [POLYGON\_ID\_FIELD], [BUFFER\_RADIUS], [OUTPUT\_POINT\_ID\_FIELD], [OUTPUT\_POLYGON\_ID\_FIELD], [BUFFER\_FIELD], [BUFFER\_VALUE], [TIE\_BREAKER], [SEGMENTS]} & For each point, create a BUFFER\_RADIUS circle clipped to the POLYGONS polygon that contains it. Output columns: OUTPUT\_POINT\_ID\_FIELD (point identifier), OUTPUT\_POLYGON\_ID\_FIELD\textbackslash dots \\
\texttt{clip\_\allowbreak geometry\_\allowbreak buffers\_\allowbreak to\_\allowbreak containing\_\allowbreak polygons}\newline {\scriptsize\itshape INPUT, CONTAINERS, OUTPUT, [INPUT\_ID\_FIELD], [CONTAINER\_ID\_FIELD], [BUFFER\_DISTANCE], [OUTPUT\_INPUT\_ID\_FIELD], [OUTPUT\_CONTAINER\_ID\_FIELD], [BUFFER\_FIELD], [AREA\_FIELD], [ROUND\_DECIMALS], [CONTAINER\_TIE\_BREAKER], [SEGMENTS]} & Buffer each INPUT geometry by BUFFER\_DISTANCE and clip the result to the CONTAINERS polygon that contains it. Output columns: OUTPUT\_INPUT\_ID\_FIELD (INPUT identifier),\textbackslash dots \\
\texttt{voronoi\_\allowbreak polygons}\newline {\scriptsize\itshape INPUT, OUTPUT, [BUFFER], [TOLERANCE], [COPY\_ATTRIBUTES]} & Generate a Voronoi diagram (Thiessen polygons) from point features, one polygon per input point. \\
\addlinespace
\multicolumn{2}{@{}l}{\textbf{Geometry construction \& conversion} (22)}\\[0.2em]
\texttt{minimum\_\allowbreak bounding\_\allowbreak geometry}\newline {\scriptsize\itshape INPUT, TYPE, OUTPUT, [FIELD]} & Create a minimum bounding geometry for each feature or, when FIELD is given, for each group of features. \\
\texttt{multipart\_\allowbreak to\_\allowbreak singleparts}\newline {\scriptsize\itshape INPUT, OUTPUT} & Explode multipart geometries into single-part features, one output feature per part with duplicated attributes. \\
\texttt{point\_\allowbreak on\_\allowbreak surface}\newline {\scriptsize\itshape INPUT, OUTPUT, [ANGLE\_TOLERANCE]} & Generate a guaranteed-interior representative point for each feature's surface. The output point is always inside or on the boundary of the feature, even for concave shapes. \\
\texttt{geometry\_\allowbreak by\_\allowbreak expression}\newline {\scriptsize\itshape INPUT, OUTPUT, EXPRESSION, [OUTPUT\_GEOMETRY], [WITH\_Z], [WITH\_M]} & Replace each feature's geometry with the result of evaluating the QGIS geometry EXPRESSION. \\
\texttt{extract\_\allowbreak specific\_\allowbreak vertices}\newline {\scriptsize\itshape INPUT, VERTICES, OUTPUT} & Extract geometry vertices at the given 0-based positions, one output point per matched vertex. \\
\texttt{features\_\allowbreak to\_\allowbreak lines}\newline {\scriptsize\itshape INPUTS, OUTPUT, [ATTRIBUTES], [CLUSTER\_TOLERANCE]} & Split line and polygon boundary geometries from INPUTS at all intersection points, producing a set of non-duplicate line segments covering all edges. \\
\texttt{polygons\_\allowbreak to\_\allowbreak lines}\newline {\scriptsize\itshape INPUT, OUTPUT, [SPLIT\_SHARED\_BOUNDARIES]} & Convert polygon boundaries to line features. When SPLIT\_SHARED\_BOUNDARIES=True, shared edges between polygons produce separate line features with LEFT\_FID (polygon on the left) and\textbackslash dots \\
\texttt{centroid\_\allowbreak points}\newline {\scriptsize\itshape INPUT, OUTPUT, [ALL\_PARTS]} & Replace each feature's geometry with its centroid point. \\
\texttt{convex\_\allowbreak hull}\newline {\scriptsize\itshape INPUT, OUTPUT} & Replace each feature's geometry with its convex hull. Output appends two columns: 'area' (polygon area in CRS units) and 'perimeter' (polygon perimeter in CRS units). \\
\texttt{extract\_\allowbreak vertices}\newline {\scriptsize\itshape INPUT, OUTPUT} & Extract every vertex of each input geometry as an individual point feature, preserving source attributes. \\
\texttt{create\_\allowbreak point\_\allowbreak layer\_\allowbreak from\_\allowbreak table}\newline {\scriptsize\itshape INPUT, XFIELD, YFIELD, TARGET\_CRS, OUTPUT, [ZFIELD], [MFIELD]} & Create a point layer from coordinate columns in a table, assigning TARGET\_CRS to the output geometries. \\
\texttt{points\_\allowbreak along\_\allowbreak lines}\newline {\scriptsize\itshape INPUT, OUTPUT, [PLACEMENT], [DISTANCE], [PERCENTAGE], [INCLUDE\_END\_POINTS], [ADD\_CHAINAGE\_FIELDS], [DISTANCE\_FIELD]} & Place points at a regular DISTANCE interval along each line. Output includes all INPUT attributes. When ADD\_CHAINAGE\_FIELDS=True, output also includes ORIG\_FID (source line identifier),\textbackslash dots \\
\texttt{explode\_\allowbreak lines}\newline {\scriptsize\itshape INPUT, OUTPUT} & Split each line at every interior node, producing individual two-vertex line segments with duplicated attributes. \\
\texttt{create\_\allowbreak grid}\newline {\scriptsize\itshape TYPE, EXTENT, HSPACING, VSPACING, CRS, OUTPUT, [HOVERLAY], [VOVERLAY]} & Create a regular vector grid covering EXTENT at HSPACING x VSPACING cell size. \\
\texttt{bounding\_\allowbreak boxes}\newline {\scriptsize\itshape INPUT, OUTPUT} & Replace each feature's geometry with its axis-aligned bounding box (minimum enclosing rectangle aligned to the CRS axes). \\
\texttt{polygonize\_\allowbreak lines}\newline {\scriptsize\itshape INPUT, OUTPUT, [KEEP\_FIELDS]} & Form polygons from closed line rings in a line network, filling every enclosed area as a polygon feature. \\
\texttt{indexed\_\allowbreak fishnet}\newline {\scriptsize\itshape TEMPLATE, CELL\_WIDTH, CELL\_HEIGHT, OUTPUT, [NUMBER\_ROWS], [NUMBER\_COLUMNS], [GEOMETRY\_TYPE], [CRS], [ROW\_FIELD], [COL\_FIELD], [ID\_FIELD], [ORIGIN\_MODE], [ROW\_ORDER]} & Create a regular grid over the extent of TEMPLATE at CELL\_WIDTH x CELL\_HEIGHT spacing. Output columns: ID\_FIELD (1-based sequential cell ID), ROW\_FIELD (row index), COL\_FIELD (column index). \\
\texttt{snap\_\allowbreak centered\_\allowbreak square\_\allowbreak boxes\_\allowbreak to\_\allowbreak raster\_\allowbreak grid}\newline {\scriptsize\itshape INPUT, RASTER, HALF\_SIZE, OUTPUT, [ID\_FIELD], [OUTPUT\_ID\_FIELD], [BUFFER\_FIELD], [BUFFER\_VALUE], [MODE\_FIELD], [MODE\_VALUE]} & Create one square polygon per input point, snapped inward to the nearest RASTER cell boundaries. The square is centered on the point with half-width HALF\_SIZE but the actual edges are\textbackslash dots \\
\texttt{add\_\allowbreak xy\_\allowbreak fields}\newline {\scriptsize\itshape INPUT, OUTPUT, [CRS], [PREFIX]} & Append X and Y coordinate fields to each feature. Output field names are PREFIX+'x' and PREFIX+'y'. \\
\texttt{centered\_\allowbreak squares\_\allowbreak from\_\allowbreak points}\newline {\scriptsize\itshape INPUT, SIDE\_LENGTHS, OUTPUT, [ID\_FIELD], [OUTPUT\_ID\_FIELD], [SIDE\_FIELD]} & Create one centered square polygon per input point per side length in SIDE\_LENGTHS; each square is centered on its source point. Output columns: OUTPUT\_ID\_FIELD (source identifier),\textbackslash dots \\
\texttt{offset\_\allowbreak points\_\allowbreak by\_\allowbreak bearings}\newline {\scriptsize\itshape INPUT, OUTPUT, DISTANCE, BEARINGS, [ORIGIN\_ID\_FIELD], [OFFSET\_INDEX\_FIELD], [BEARING\_FIELD], [DISTANCE\_FIELD]} & Create one offset point per input point per bearing, translating each point by DISTANCE in the given direction. Output columns: ORIGIN\_ID\_FIELD (source feature identifier),\textbackslash dots \\
\texttt{translate\_\allowbreak points\_\allowbreak by\_\allowbreak xy\_\allowbreak offset}\newline {\scriptsize\itshape INPUT, DX, DY, OUTPUT, [ID\_FIELD], [OUTPUT\_ID\_FIELD], [X\_FIELD], [Y\_FIELD], [ROUND\_DECIMALS]} & Shift every point geometry by (DX, DY) in CRS units, appending OUTPUT\_ID\_FIELD, X\_FIELD (new X, rounded to ROUND\_DECIMALS), and Y\_FIELD (new Y, rounded to ROUND\_DECIMALS). \\
\addlinespace
\multicolumn{2}{@{}l}{\textbf{Geometry cleaning \& dissolve} (7)}\\[0.2em]
\texttt{delete\_\allowbreak duplicates\_\allowbreak by\_\allowbreak attribute}\newline {\scriptsize\itshape INPUT, FIELDS, OUTPUT, [DUPLICATES]} & Remove duplicate features by matching values in FIELDS, keeping the first occurrence; optionally write removed duplicates to DUPLICATES. \\
\texttt{find\_\allowbreak identical\_\allowbreak features}\newline {\scriptsize\itshape INPUT, FIELDS, OUTPUT, [XY\_TOLERANCE], [Z\_TOLERANCE], [OUTPUT\_RECORD\_OPTION]} & Find features with identical attribute values or geometry and write a table of duplicates. Output columns: IN\_FID (1-based feature ID), FEAT\_SEQ (group identifier; features in the same\textbackslash dots \\
\texttt{delete\_\allowbreak duplicate\_\allowbreak geometries}\newline {\scriptsize\itshape INPUT, OUTPUT} & Remove features whose geometry is exactly duplicated by another feature, keeping the first occurrence. \\
\texttt{dissolve\_\allowbreak features}\newline {\scriptsize\itshape INPUT, OUTPUT, [FIELD], [SEPARATE\_DISJOINT]} & Merge features into multipart geometries by grouping on FIELD values. All features merge into one when FIELD is empty. \\
\texttt{fix\_\allowbreak geometries}\newline {\scriptsize\itshape INPUT, OUTPUT, [METHOD]} & Repair topologically invalid vector geometries using GEOS algorithms. \\
\texttt{delete\_\allowbreak holes}\newline {\scriptsize\itshape INPUT, OUTPUT, [MIN\_AREA]} & Remove interior rings (holes) from polygon features. Holes with area strictly less than MIN\_AREA are removed; MIN\_AREA=0.0 keeps all holes; pass a large value (e.g., float('inf')) to\textbackslash dots \\
\texttt{simplify\_\allowbreak geometries}\newline {\scriptsize\itshape INPUT, METHOD, TOLERANCE, OUTPUT} & Reduce vertex count of geometries using a named simplification algorithm. \\
\addlinespace
\multicolumn{2}{@{}l}{\textbf{Attribute \& table operations} (15)}\\[0.2em]
\texttt{calculate\_\allowbreak attribute\_\allowbreak field}\newline {\scriptsize\itshape INPUT, FIELD\_NAME, FORMULA, [OUTPUT], [FIELD\_TYPE], [FIELD\_LENGTH], [FIELD\_PRECISION], [NEW\_FIELD]} & Add or overwrite attribute field FIELD\_NAME on each feature using a QGIS field-calculator FORMULA expression. \\
\texttt{join\_\allowbreak table\_\allowbreak by\_\allowbreak field}\newline {\scriptsize\itshape INPUT, FIELD, INPUT\_2, FIELD\_2, [OUTPUT], [FIELDS\_TO\_COPY], [METHOD], [DISCARD\_NONMATCHING], [PREFIX]} & Join INPUT\_2 attributes onto INPUT by matching FIELD = FIELD\_2, retaining INPUT geometry. Accepts '\$id' as either FIELD or FIELD\_2. \\
\texttt{extract\_\allowbreak features\_\allowbreak by\_\allowbreak expression}\newline {\scriptsize\itshape INPUT, EXPRESSION, OUTPUT, [FAIL\_OUTPUT]} & Extract features where the QGIS expression EXPRESSION evaluates to true, writing matched features to OUTPUT. \\
\texttt{sort\_\allowbreak table\_\allowbreak rows}\newline {\scriptsize\itshape INPUT, SORT\_FIELDS, OUTPUT} & Sort rows by one or more fields. SORT\_FIELDS are applied in order, so the first entry is the primary sort key. \\
\texttt{statistics\_\allowbreak table}\newline {\scriptsize\itshape INPUT, AGGREGATES, OUTPUT, [CASE\_FIELDS]} & Aggregate INPUT rows using AGGREGATES function definitions, optionally grouped by CASE\_FIELDS. Output columns: CASE\_FIELDS columns (one per group-by field) plus one column per AGGREGATES\textbackslash dots \\
\texttt{add\_\allowbreak geometry\_\allowbreak attributes}\newline {\scriptsize\itshape INPUT, [OUTPUT], [GEOMETRY\_PROPERTIES], [AREA\_UNIT], [LENGTH\_UNIT]} & Append geometric measurement columns to each feature. \\
\texttt{rename\_\allowbreak field}\newline {\scriptsize\itshape INPUT, FIELD, NEW\_NAME, OUTPUT} & Rename attribute field FIELD to NEW\_NAME in the output layer. The QGIS expression '\$id' is accepted for FIELD, which materializes the feature ID as a real column named NEW\_NAME. \\
\texttt{add\_\allowbreak attribute\_\allowbreak field}\newline {\scriptsize\itshape INPUT, FIELD\_NAME, FIELD\_TYPE, OUTPUT, [FIELD\_LENGTH], [FIELD\_PRECISION]} & Add an empty attribute field FIELD\_NAME of the given type to a vector layer, initialized to NULL for all features. \\
\texttt{retain\_\allowbreak fields}\newline {\scriptsize\itshape INPUT, FIELDS, OUTPUT} & Keep only the listed FIELDS attribute columns, dropping all others. Geometry is always retained. An empty FIELDS list drops all attributes. \\
\texttt{delete\_\allowbreak columns}\newline {\scriptsize\itshape INPUT, COLUMN, OUTPUT} & Remove the listed COLUMN attribute fields from the layer, keeping all other fields and geometry. COLUMN accepts a list of field name strings. \\
\texttt{statistics\_\allowbreak by\_\allowbreak categories}\newline {\scriptsize\itshape INPUT, CATEGORIES\_FIELD\_NAME, OUTPUT, [VALUES\_FIELD\_NAME]} & Compute descriptive statistics for VALUES\_FIELD\_NAME grouped by unique combinations of CATEGORIES\_FIELD\_NAME. Output columns: CATEGORIES\_FIELD\_NAME fields plus count, unique, min, max,\textbackslash dots \\
\texttt{join\_\allowbreak attributes\_\allowbreak by\_\allowbreak field\_\allowbreak value}\newline {\scriptsize\itshape INPUT, FIELD, INPUT\_2, FIELD\_2, OUTPUT, [FIELDS\_TO\_COPY], [METHOD], [DISCARD\_NONMATCHING], [PREFIX], [NON\_MATCHING]} & Join INPUT\_2 attributes onto INPUT by matching FIELD = FIELD\_2, retaining INPUT geometry. \\
\texttt{frequency\_\allowbreak table}\newline {\scriptsize\itshape INPUT, FIELDS, OUTPUT, [COUNT\_FIELD]} & Count the occurrences of each unique combination of FIELDS values. Output columns: FIELDS columns plus COUNT\_FIELD (integer occurrence count). \\
\texttt{wide\_\allowbreak table\_\allowbreak to\_\allowbreak long\_\allowbreak table}\newline {\scriptsize\itshape INPUT, ID\_FIELDS, [VARIABLE\_FIELD], [VALUE\_FIELD], [OUTPUT], [WIDE\_FIELDS], [WIDE\_FIELD\_PREFIX], [ID\_OUTPUT\_FIELDS], [INCLUDE\_ZERO], [VARIABLE\_FIELD\_TYPE], [VALUE\_FIELD\_TYPE]} & Reshape a wide table to long format. Output columns: ID\_OUTPUT\_FIELDS (one per ID\_FIELDS entry), VARIABLE\_FIELD (column name stripped of prefix), VALUE\_FIELD (cell value). \\
\texttt{add\_\allowbreak autoincremental\_\allowbreak field}\newline {\scriptsize\itshape INPUT, FIELD\_NAME, OUTPUT, [START], [MODULUS], [GROUP\_FIELDS], [SORT\_EXPRESSION], [SORT\_ASCENDING], [SORT\_NULLS\_FIRST]} & Add an integer FIELD\_NAME that increments by 1 for each feature starting at START (default 0, so the first feature receives 0). \\
\addlinespace
\multicolumn{2}{@{}l}{\textbf{Spatial join \& selection} (6)}\\[0.2em]
\texttt{join\_\allowbreak attributes\_\allowbreak by\_\allowbreak location}\newline {\scriptsize\itshape INPUT, JOIN, OUTPUT, [PREDICATE\_CODES], [METHOD], [DISCARD\_NONMATCHING], [JOIN\_FIELDS], [PREFIX], [ADD\_TARGET\_FID], [ADD\_JOIN\_COUNT]} & Join JOIN attributes onto INPUT features that satisfy the spatial predicates. Optionally appends Join\_Count (number of matching JOIN features per INPUT feature). \\
\texttt{join\_\allowbreak attributes\_\allowbreak by\_\allowbreak location\_\allowbreak summary}\newline {\scriptsize\itshape INPUT, JOIN, OUTPUT, [PREDICATE\_CODES], [JOIN\_FIELDS], [SUMMARIES], [DISCARD\_NONMATCHING], [PREFIX]} & Aggregate JOIN feature attributes into per-INPUT-feature summary statistics using spatial relationship. Output appends '<field>\_<stat>' columns for each JOIN\_FIELDSxSUMMARIES combination. \\
\texttt{extract\_\allowbreak features\_\allowbreak by\_\allowbreak location}\newline {\scriptsize\itshape INPUT, INTERSECT, PREDICATE, OUTPUT} & Extract INPUT features that satisfy at least one spatial predicate against the INTERSECT layer. \\
\texttt{point\_\allowbreak polygon\_\allowbreak layer\_\allowbreak counts}\newline {\scriptsize\itshape POINTS, POLYGON\_LAYERS, OUTPUT, [POINT\_ID\_FIELD], [OUTPUT\_COUNT\_FIELDS], [PREDICATE], [SORT\_BY\_ID]} & Count the number of polygon features from each POLYGON\_LAYERS layer that satisfy PREDICATE for each point. Output columns: POINT\_ID\_FIELD plus one count column per layer in\textbackslash dots \\
\texttt{join\_\allowbreak attributes\_\allowbreak by\_\allowbreak nearest}\newline {\scriptsize\itshape INPUT, INPUT\_2, OUTPUT, [FIELDS\_TO\_COPY], [DISCARD\_NONMATCHING], [PREFIX], [NEIGHBORS], [MAX\_DISTANCE], [ADD\_TARGET\_FID]} & Join INPUT\_2 attributes onto each INPUT feature using nearest-neighbor geometry matching, retaining INPUT geometry. \\
\texttt{join\_\allowbreak attributes\_\allowbreak by\_\allowbreak location\_\allowbreak within\_\allowbreak distance}\newline {\scriptsize\itshape INPUT, JOIN, DISTANCE, OUTPUT, [METHOD], [DISCARD\_NONMATCHING], [JOIN\_FIELDS], [PREFIX], [ADD\_JOIN\_COUNT]} & Join JOIN attributes onto INPUT features whose geometries lie within planar DISTANCE, preserving INPUT geometry. \\
\addlinespace
\multicolumn{2}{@{}l}{\textbf{Raster processing} (15)}\\[0.2em]
\texttt{clip\_\allowbreak raster\_\allowbreak by\_\allowbreak mask}\newline {\scriptsize\itshape INPUT, MASK, OUTPUT, [SOURCE\_CRS], [TARGET\_CRS], [NODATA], [ALPHA\_BAND], [CROP\_TO\_CUTLINE], [KEEP\_RESOLUTION], [SET\_RESOLUTION], [X\_RESOLUTION], [Y\_RESOLUTION], [MULTITHREADING], [OPTIONS], [DATA\_TYPE], [EXTRA]} & Clip a raster to the shape of vector MASK polygons using GDAL. \\
\texttt{clip\_\allowbreak raster\_\allowbreak to\_\allowbreak raster\_\allowbreak extent}\newline {\scriptsize\itshape INPUT, REFERENCE\_RASTER, OUTPUT, [NODATA], [DATA\_TYPE], [EXTRA]} & Clip INPUT to the axis-aligned bounding box of REFERENCE\_RASTER. \\
\texttt{clip\_\allowbreak raster\_\allowbreak to\_\allowbreak layer\_\allowbreak extent}\newline {\scriptsize\itshape INPUT, REFERENCE\_LAYER, OUTPUT, [NODATA], [DATA\_TYPE], [EXTRA]} & Clip INPUT to the axis-aligned bounding box of REFERENCE\_LAYER (vector or raster). \\
\texttt{clump\_\allowbreak raster\_\allowbreak regions}\newline {\scriptsize\itshape INPUT, OUTPUT, [DIAGONAL], [THRESHOLD], [TITLE], [GRASS\_REGION\_PARAMETER], [GRASS\_REGION\_CELLSIZE\_PARAMETER]} & Label connected raster regions of equal pixel value with a unique integer component ID. Output is an Int32 raster where each distinct contiguous region has a unique positive ID; 0\textbackslash dots \\
\texttt{rearrange\_\allowbreak raster\_\allowbreak bands}\newline {\scriptsize\itshape INPUT, BANDS, OUTPUT, [DATA\_TYPE], [CREATION\_OPTIONS]} & Reorder or subset raster bands; output has one band per entry in BANDS (1-based band indices). \\
\texttt{raster\_\allowbreak calculator}\newline {\scriptsize\itshape LAYERS, FORMULA, OUTPUT, [RTYPE], [NO\_DATA], [EXTENT\_OPT], [EXTRA]} & Evaluate a pixel-wise arithmetic FORMULA across LAYERS, where each layer is referenced as A, B, C, ... in LAYERS order (up to six layers). \\
\texttt{focal\_\allowbreak statistics\_\allowbreak raster}\newline {\scriptsize\itshape INPUT, OUTPUT, [METHOD], [CIRCULAR], [RADIUS\_OR\_WIDTH], [HEIGHT], [UNITS], [IGNORE\_NODATA], [WINDOW\_MODE]} & Compute a per-cell statistic over a moving neighborhood window, producing a Float32 output raster. \\
\texttt{reclassify\_\allowbreak raster\_\allowbreak by\_\allowbreak table}\newline {\scriptsize\itshape INPUT\_RASTER, TABLE, OUTPUT, [RASTER\_BAND], [NO\_DATA], [RANGE\_BOUNDARIES], [NODATA\_FOR\_MISSING], [DATA\_TYPE]} & Reclassify raster values using a TABLE of [lower\_bound, upper\_bound, new\_value] rows. Each row maps cells whose values fall in the specified range to new\_value. \\
\texttt{cell\_\allowbreak statistics\_\allowbreak raster}\newline {\scriptsize\itshape INPUT, STATISTIC, REF\_LAYER, OUTPUT, [IGNORE\_NODATA], [OUTPUT\_NO\_DATA\_VALUE], [CREATION\_OPTIONS]} & Compute a cell-by-cell statistic across a stack of aligned rasters, using REF\_LAYER to define the output grid. All INPUT rasters must share the same CRS, origin, and cell size as\textbackslash dots \\
\texttt{merge\_\allowbreak rasters}\newline {\scriptsize\itshape INPUT, OUTPUT, [PCT], [SEPARATE], [NODATA\_INPUT], [NODATA\_OUTPUT], [DATA\_TYPE], [OPTIONS], [EXTRA]} & Mosaic multiple rasters into a single-band output by overlaying them in order (last INPUT on top). \\
\texttt{region\_\allowbreak group\_\allowbreak raster}\newline {\scriptsize\itshape INPUT, OUTPUT, [NEIGHBORS], [EXCLUDED\_VALUE]} & Label connected regions of equal-value raster cells with unique positive integers. Produces an Int32 raster; 0 marks NoData or EXCLUDED\_VALUE cells. NEIGHBORS must be 4 or 8;\textbackslash dots \\
\texttt{aggregate\_\allowbreak raster}\newline {\scriptsize\itshape INPUT, CELL\_FACTOR, STATISTIC, OUTPUT, [IGNORE\_NODATA]} & Reduce raster resolution by aggregating CELL\_FACTOR x CELL\_FACTOR blocks into a single output cell using STATISTIC. The output grid origin aligns with the input origin; each output cell\textbackslash dots \\
\texttt{translate\_\allowbreak raster\_\allowbreak format}\newline {\scriptsize\itshape INPUT, OUTPUT, [NODATA], [COPY\_SUBDATASETS], [CREATION\_OPTIONS], [EXTRA], [DATA\_TYPE]} & Convert or copy a raster to a new file, optionally changing pixel data type. \\
\texttt{raster\_\allowbreak value\_\allowbreak count\_\allowbreak table}\newline {\scriptsize\itshape INPUT\_RASTER, OUTPUT} & Count pixel occurrences for each unique integer value in band 1 of INPUT\_RASTER. Output columns: value (pixel value), count (number of cells with that value). \\
\texttt{weighted\_\allowbreak sum\_\allowbreak raster}\newline {\scriptsize\itshape INPUTS, WEIGHTS, OUTPUT} & Compute a pixel-wise weighted sum of already aligned rasters. Each INPUTS[i] is multiplied by WEIGHTS[i] and the products are summed. All inputs must share the same CRS, transform, and\textbackslash dots \\
\addlinespace
\multicolumn{2}{@{}l}{\textbf{Raster--vector conversion \& sampling} (11)}\\[0.2em]
\texttt{polygonize\_\allowbreak raster}\newline {\scriptsize\itshape INPUT, OUTPUT, [BAND], [FIELD], [EIGHT\_CONNECTEDNESS], [EXTRA]} & Convert contiguous raster regions of equal pixel value in BAND into polygon features, storing the pixel value in FIELD. \\
\texttt{raster\_\allowbreak max\_\allowbreak points\_\allowbreak by\_\allowbreak polygon}\newline {\scriptsize\itshape POLYGONS, RASTER, OUTPUT, [POLYGON\_ID\_FIELD], [OUTPUT\_ID\_FIELD], [OUTPUT\_VALUE\_FIELD], [ROUND\_DECIMALS], [INCLUDE\_BOUNDARY]} & Create one point feature per polygon at the raster cell center with the maximum pixel value inside that polygon. Output columns: OUTPUT\_ID\_FIELD (polygon identifier from\textbackslash dots \\
\texttt{raster\_\allowbreak pixels\_\allowbreak to\_\allowbreak points}\newline {\scriptsize\itshape INPUT\_RASTER, RASTER\_BAND, FIELD\_NAME, OUTPUT} & Create one point at each valid raster cell center for RASTER\_BAND, storing the pixel value in FIELD\_NAME. NoData cells produce no point. \\
\texttt{sample\_\allowbreak raster\_\allowbreak values\_\allowbreak to\_\allowbreak points}\newline {\scriptsize\itshape INPUT, RASTER, COLUMN\_PREFIX, OUTPUT} & Sample RASTER cell values at each point location, appending columns COLUMN\_PREFIX1, COLUMN\_PREFIX2, ... (one per band). NoData cells produce NULL. \\
\texttt{raster\_\allowbreak domain\_\allowbreak polygon}\newline {\scriptsize\itshape INPUT\_RASTER, OUTPUT, [BAND]} & Create polygon features covering all contiguous regions of valid (non-NoData, finite) cells in BAND. Multiple disconnected regions produce multiple features; output column gridcode=1 for\textbackslash dots \\
\texttt{polygon\_\allowbreak to\_\allowbreak raster}\newline {\scriptsize\itshape INPUT, FIELD, OUTPUT, CELL\_SIZE, [ALL\_TOUCHED], [PRIORITY\_FIELD], [DATA\_TYPE], [EXTENT], [INIT], [NODATA], [EXTRA], [EXTENT\_BUFFER]} & Burn vector polygon attribute FIELD into a raster grid at CELL\_SIZE resolution. \\
\texttt{point\_\allowbreak to\_\allowbreak raster}\newline {\scriptsize\itshape INPUT, FIELD, OUTPUT, CELL\_SIZE, [DATA\_TYPE], [EXTENT], [INIT], [NODATA], [EXTRA]} & Burn point feature attribute FIELD into a raster grid at CELL\_SIZE resolution. \\
\texttt{polyline\_\allowbreak to\_\allowbreak raster}\newline {\scriptsize\itshape INPUT, FIELD, OUTPUT, CELL\_SIZE, [ALL\_TOUCHED], [PRIORITY\_FIELD], [DATA\_TYPE], [EXTENT], [INIT], [NODATA], [EXTRA]} & Burn line feature attribute FIELD into a raster grid at CELL\_SIZE resolution. \\
\texttt{sample\_\allowbreak multiple\_\allowbreak rasters\_\allowbreak at\_\allowbreak points}\newline {\scriptsize\itshape RASTERS, LOCATIONS, OUTPUT, [COLUMN\_PREFIX]} & Sample values from each raster in RASTERS at LOCATIONS points, merging RASTERS into a multi-band stack first. Output columns from LOCATIONS plus COLUMN\_PREFIX1, COLUMN\_PREFIX2, ... (one\textbackslash dots \\
\texttt{raster\_\allowbreak contour\_\allowbreak lines}\newline {\scriptsize\itshape INPUT, OUTPUT, [BAND], [INTERVAL], [FIELD\_NAME], [CREATE\_3D], [IGNORE\_NODATA], [NODATA], [OFFSET], [OPTIONS], [EXTRA]} & Generate contour lines from raster BAND at a fixed elevation INTERVAL. Output column FIELD\_NAME stores the contour elevation value for each line. \\
\texttt{raster\_\allowbreak cells\_\allowbreak to\_\allowbreak polygons}\newline {\scriptsize\itshape INPUT\_RASTER, OUTPUT, [INCLUDE\_NODATA], [ROW\_FIELD], [COL\_FIELD], [CELL\_ID\_FIELD], [VALUE\_FIELD]} & Create one polygon per raster cell, one feature per valid pixel. Output columns: ROW\_FIELD (row index), COL\_FIELD (column index), CELL\_ID\_FIELD (sequential cell ID), VALUE\_FIELD (pixel\textbackslash dots \\
\addlinespace
\multicolumn{2}{@{}l}{\textbf{Zonal statistics \& area tabulation} (5)}\\[0.2em]
\texttt{tabulate\_\allowbreak area\_\allowbreak table}\newline {\scriptsize\itshape ZONES, ZONE\_FIELD, CLASSES, [CLASS\_FIELD], [PROCESSING\_CELL\_SIZE], [OUTPUT]} & Count cells of each CLASSES raster value within each ZONES polygon and multiply by pixel area. Output columns: ZONE\_FIELD (zone identifier) plus one VALUE\_<n> column per raster class\textbackslash dots \\
\texttt{zonal\_\allowbreak statistics\_\allowbreak as\_\allowbreak table}\newline {\scriptsize\itshape ZONES, ZONE\_FIELD, RASTER, STATISTICS, COLUMN\_PREFIX, [FIELD\_RENAMES], [IGNORE\_NODATA], [OUTPUT]} & Compute raster statistics for each polygon zone and write them as a table (geometry dropped). Output columns: all ZONES attribute columns plus one per STATISTICS entry named\textbackslash dots \\
\texttt{zonal\_\allowbreak statistics\_\allowbreak raster}\newline {\scriptsize\itshape ZONES, ZONE\_FIELD, RASTER, [STATISTIC\_FIELD], [IGNORE\_NODATA], [OUTPUT]} & Compute zonal statistics of RASTER by ZONES and burn the STATISTIC\_FIELD value back onto a raster grid matching RASTER. STATISTIC\_FIELD must be MAX or COUNT (from zonal\_statistics\_as\_table). \\
\texttt{tabulate\_\allowbreak intersection}\newline {\scriptsize\itshape ZONES, ZONE\_FIELDS, CLASSES, [CLASS\_FIELDS], [SUM\_FIELDS], [OUT\_UNITS], [OUTPUT]} & Intersect ZONES and CLASSES polygon layers and compute overlap areas. Output columns: ZONE\_FIELDS columns, CLASS\_FIELDS columns, AREA (intersection area in the squared CRS units of\textbackslash dots \\
\texttt{zonal\_\allowbreak percentile\_\allowbreak table}\newline {\scriptsize\itshape ZONES, ZONE\_FIELD, RASTER, [PERCENTILE], [INTERPOLATION], [IGNORE\_NODATA], [OUTPUT]} & Compute a raster percentile for each polygon zone and write a table. Output columns: ZONE\_FIELD (zone identifier), COUNT (valid cell count), AREA (COUNT x pixel area in CRS units), PCTxx\textbackslash dots \\
\addlinespace
\multicolumn{2}{@{}l}{\textbf{Density surfaces} (3)}\\[0.2em]
\texttt{gaussian\_\allowbreak kernel\_\allowbreak from\_\allowbreak points}\newline {\scriptsize\itshape POINTS, TEMPLATE\_RASTER, OUTPUT, SIGMA, [NODATA], [ROUND\_DECIMALS]} & Create a raster where each valid cell value is exp(-d\textasciicircum 2 / (2*sigma\textasciicircum 2)) where d is the Euclidean distance to the nearest POINTS location. Output uses the same grid as TEMPLATE\_RASTER. \\
\texttt{heatmap\_\allowbreak kernel\_\allowbreak density}\newline {\scriptsize\itshape INPUT, RADIUS, PIXEL\_SIZE, OUTPUT, [WEIGHT\_FIELD], [RADIUS\_FIELD], [KERNEL], [DECAY], [OUTPUT\_VALUE]} & Create a kernel density estimation raster from point features; each cell value is the density contribution within RADIUS of the cell center. \\
\texttt{point\_\allowbreak density\_\allowbreak raster}\newline {\scriptsize\itshape point\_fc, output\_path, [population\_field], [cell\_size], [neighborhood\_type], [neighborhood\_radius], [neighborhood\_units], [area\_unit\_scale\_factor]} & Create a kernel density estimation raster from point features. \\
\addlinespace
\multicolumn{2}{@{}l}{\textbf{Terrain \& hydrology} (9)}\\[0.2em]
\texttt{slope\_\allowbreak raster}\newline {\scriptsize\itshape INPUT, Z\_FACTOR, OUTPUT} & Compute a slope raster in degrees from an elevation model, using Z\_FACTOR to scale vertical units relative to horizontal units. \\
\texttt{aspect\_\allowbreak raster}\newline {\scriptsize\itshape INPUT, OUTPUT, [Z\_FACTOR]} & Compute an aspect raster (compass direction of steepest descent in degrees) from an elevation model. Z\_FACTOR scales vertical units relative to horizontal units. \\
\texttt{viewshed\_\allowbreak raster}\newline {\scriptsize\itshape raster\_path, observer\_features, output\_path, [z\_factor], [curvature\_correction], [refractivity\_coefficient]} & Create a binary union viewshed from observer point features. Output is a raster where each cell is 1 if visible from any observer and 0 otherwise. \\
\texttt{hillshade\_\allowbreak raster}\newline {\scriptsize\itshape INPUT, OUTPUT, [BAND], [Z\_FACTOR], [SCALE], [AZIMUTH], [ALTITUDE], [COMPUTE\_EDGES], [ZEVENBERGEN], [COMBINED], [MULTIDIRECTIONAL], [EXTRA], [CREATION\_OPTIONS]} & Generate a hillshade raster from an elevation model. Z\_FACTOR exaggerates vertical scale; AZIMUTH is the light-source compass angle (0=north); ALTITUDE is the sun elevation angle in\textbackslash dots \\
\texttt{fill\_\allowbreak sinks}\newline {\scriptsize\itshape raster\_path, output\_path, [z\_limit]} & Fill topographic sinks in a DEM to produce continuous flow paths. z\_limit: maximum depth of sinks to fill; pass None or 0 for unlimited. \\
\texttt{flow\_\allowbreak direction\_\allowbreak raster}\newline {\scriptsize\itshape surface\_raster, output\_path, [force\_flow], [flow\_direction\_type]} & Compute the downslope flow direction for each cell of a sink-filled DEM. force\_flow must be NORMAL; flow\_direction\_type must be D8. \\
\texttt{flow\_\allowbreak accumulation\_\allowbreak raster}\newline {\scriptsize\itshape flow\_direction\_raster\_path, output\_path, [data\_type], [flow\_direction\_type]} & Compute the number of upstream cells draining into each cell from a D8 flow-direction raster. flow\_direction\_type must be D8; data\_type should be FLOAT32 or FLOAT64 (passing FLOAT uses\textbackslash dots \\
\texttt{snap\_\allowbreak pour\_\allowbreak point\_\allowbreak raster}\newline {\scriptsize\itshape pour\_point\_data, accumulation\_raster, output\_path, snap\_distance, pour\_point\_field} & Rasterize pour-point features onto a grid matching accumulation\_raster, burning pour\_point\_field values into a Float32 raster aligned to the accumulation grid. Cells with no pour point\textbackslash dots \\
\texttt{watershed\_\allowbreak raster}\newline {\scriptsize\itshape flow\_direction\_raster\_path, pour\_point\_data, output\_path, [pour\_point\_field]} & Delineate drainage basins by assigning each valid cell to the nearest pour-point label. pour\_point\_data must be a rasterized pour-point raster (from snap\_pour\_point\_raster) aligned to\textbackslash dots \\
\addlinespace
\multicolumn{2}{@{}l}{\textbf{Network analysis \& linear referencing} (4)}\\[0.2em]
\texttt{create\_\allowbreak routes}\newline {\scriptsize\itshape INPUT, ROUTE\_ID\_FIELD, OUTPUT, [MEASURE\_FACTOR], [MEASURE\_OFFSET]} & Prepare INPUT line features as routes for linear referencing. Measures are derived from cumulative line length; MEASURE\_FACTOR multiplies and MEASURE\_OFFSET shifts all measure values.\textbackslash dots \\
\texttt{locate\_\allowbreak features\_\allowbreak along\_\allowbreak routes}\newline {\scriptsize\itshape FEATURES, ROUTES, ROUTE\_ID\_FIELD, TOLERANCE, OUTPUT, [DISTANCE\_FIELD], [ROUTE\_LOCATIONS], [INPUT\_ID\_FIELD]} & Project FEATURES onto ROUTES by linear referencing. Output columns: INPUTOID (FEATURES identifier), ROUTE\_ID\_FIELD (matched route identifier), MEAS (linear distance along route from\textbackslash dots \\
\texttt{service\_\allowbreak area\_\allowbreak from\_\allowbreak layer}\newline {\scriptsize\itshape NETWORK, START\_POINTS, TRAVEL\_COST, OUTPUT, [STRATEGY], [DIRECTION\_FIELD], [VALUE\_FORWARD], [VALUE\_BACKWARD], [VALUE\_BOTH], [SPEED\_FIELD], [DEFAULT\_SPEED], [TOLERANCE]} & Return network edges reachable within TRAVEL\_COST (meters for SHORTEST, minutes for FASTEST) from each START\_POINT. \\
\texttt{shortest\_\allowbreak path\_\allowbreak point\_\allowbreak to\_\allowbreak layer}\newline {\scriptsize\itshape NETWORK, START\_POINTS, END\_POINTS, OUTPUT, [STRATEGY], [DIRECTION\_FIELD], [VALUE\_FORWARD], [VALUE\_BACKWARD], [VALUE\_BOTH], [SPEED\_FIELD], [DEFAULT\_SPEED], [TOLERANCE]} & Compute the shortest/fastest network path from each START\_POINT to its nearest END\_POINT. \\
\addlinespace
\multicolumn{2}{@{}l}{\textbf{Input/output \& layer management} (3)}\\[0.2em]
\texttt{export\_\allowbreak table\_\allowbreak to\_\allowbreak csv}\newline {\scriptsize\itshape INPUT, OUTPUT, [FIELDS], [FIELD\_RENAMES], [EXPRESSION], [MAX\_ROWS]} & Write a table or vector layer to a CSV file at OUTPUT, dropping geometry. \\
\texttt{save\_\allowbreak features}\newline {\scriptsize\itshape INPUT, OUTPUT, [LAYER\_NAME], [DATASOURCE\_OPTIONS], [LAYER\_OPTIONS], [ACTION\_ON\_EXISTING\_FILE]} & Write a vector layer to a file, applying OGR format options and controlling existing-file behavior. \\
\texttt{merge\_\allowbreak vector\_\allowbreak layers}\newline {\scriptsize\itshape LAYERS, OUTPUT, [CRS], [ADD\_SOURCE\_FIELDS]} & Stack multiple vector layers into a single output layer, combining all features. \\
\addlinespace
\end{longtable}
}
\twocolumn

\section{Area-of-Interest Data}
\label{app:aoi}

\begin{table*}[htbp]
\centering\small
\setlength{\tabcolsep}{4pt}
\caption{Areas of interest: landscape character, analysis CRS, and task counts by family. Analysis CRS is the projected reference in which measurements are defined; several source products ship in geographic coordinates, which is one origin of the CRS Misalignment caveat.}
\label{tab:aoi_full}
\begin{tabular}{l L{4.3cm} c ccccc c}
\toprule
\textbf{Area of interest} & \textbf{Landscape} & \textbf{Analysis CRS} & \textbf{SJ\&G} & \textbf{SO\&SA} & \textbf{SPA} & \textbf{TM\&HA} & \textbf{RS\&IA} & \textbf{Tasks} \\
\midrule
San Francisco Bay Area & Dense urban coastal; bay shoreline, terrain, imagery & EPSG:26910 & 28 & 31 & 14 & 19 & 7 & 99 \\
Randstad, Netherlands  & Low-lying delta; nationally standardized registries  & EPSG:28992 & 27 & 23 & 10 &  4 & 27 & 91 \\
New York City          & Dense urban; parcel- and address-rich cadastre       & EPSG:2263  & 18 & 17 & 10 &  4 &  2 & 51 \\
Denver Front Range     & Mountain front; wildland--urban interface, hazards   & EPSG:26913 &  6 &  5 &  6 & 16 &  7 & 40 \\
Houston--Harris        & Flat coastal plain; floodplain and drainage network  & EPSG:2278  &  4 & 12 &  5 & 17 &  2 & 40 \\
South Florida          & Engineered water control; low-relief coastal wetland & EPSG:2236  & 13 &  5 &  7 &  3 &  0 & 28 \\
\midrule
\textbf{Total} & & & \textbf{96} & \textbf{93} & \textbf{52} & \textbf{63} & \textbf{45} & \textbf{349} \\
\bottomrule
\end{tabular}
\end{table*}

Table~\ref{tab:aoi_layers} reports the principal vector layers per area of interest with exact feature counts, and Table~\ref{tab:aoi_rasters} reports every raster with its dimensions and spatial reference. Feature counts are measured from the shipped files rather than declared in metadata.

Note that several rasters ship in geographic coordinate systems while the analysis reference for their area of interest is projected, and that filenames encoding a nominal resolution do not always match the shipped grid. Both conditions are genuine properties of the released data and are among the sources of the CRS Misalignment and Unit Mismatch caveats.

\begin{table}[htbp]
\centering\small
\setlength{\tabcolsep}{4pt}
\caption{Principal vector layers by area of interest, with exact feature counts.}
\label{tab:aoi_layers}
\begin{tabular}{@{}llr@{}}
\toprule
\textbf{Area of interest} & \textbf{Layer} & \textbf{Features} \\
\midrule
\multirow{4}{*}{San Francisco Bay}
 & police\_incidents      & 494{,}975 \\
 & parcels                & 236{,}300 \\
 & buildings              & 177{,}023 \\
 & roads\_osm             & 27{,}605 \\
\addlinespace
\multirow{4}{*}{New York City}
 & address\_point         & 500{,}000 \\
 & buildings              & 142{,}257 \\
 & pluto\_lots            & 133{,}289 \\
 & lion (street network)  & 66{,}234 \\
\addlinespace
\multirow{3}{*}{South Florida}
 & address\_points        & 939{,}821 \\
 & miami\_parcels         & 592{,}689 \\
 & street\_network        & 115{,}981 \\
\addlinespace
\multirow{3}{*}{Houston--Harris}
 & harris\_roads          & 365{,}800 \\
 & hcad\_parcels\_clean   & 312{,}518 \\
 & canals                 & 3{,}957 \\
\addlinespace
\multirow{3}{*}{Denver Front Range}
 & building\_footprints   & 988{,}345 \\
 & census\_tracts         & 324 \\
 & lookouts               & 208 \\
\addlinespace
\multirow{3}{*}{Randstad}
 & roads\_osm             & 55{,}346 \\
 & address\_points        & 5{,}500 \\
 & cbs\_buurten\_2025     & 1{,}000 \\
\bottomrule
\end{tabular}
\end{table}

\begin{table}[htbp]
\centering\small
\setlength{\tabcolsep}{3pt}
\caption{Raster products by area of interest.}
\label{tab:aoi_rasters}
\begin{tabular}{@{}llrl@{}}
\toprule
\textbf{Area} & \textbf{Raster} & \textbf{Pixels} & \textbf{CRS} \\
\midrule
Denver & dem\_81m                & $1164\times889$   & 4258 \\
Denver & friction\_surface\_01   & $1164\times889$   & 4258 \\
Denver & friction\_surface\_02   & $1164\times889$   & 4258 \\
Denver & wildfire\_risk\_2024    & $1152\times886$   & 4269 \\
Denver & topo\_scan\_1950        & $1152\times886$   & 4326 \\
Houston & dem\_34m               & $3663\times2428$  & 4258 \\
Houston & naip\_latest           & $4096\times3732$  & 2278 \\
NYC & dem\_10m                   & $5345\times5287$  & 2263 \\
NYC & dem\_1m                    & $4221\times2871$  & 4326 \\
NYC & nypl\_map\_scan\_1924      & $11266\times7777$ & 4326 \\
NYC & nypl\_map\_scan\_1951      & $4560\times5657$  & 4326 \\
NYC & naip\_latest               & $4045\times4090$  & 2263 \\
SF Bay & dem\_1m                 & $1817\times1390$  & 4269 \\
SF Bay & naip\_latest            & $4096\times3841$  & 26910 \\
S. Florida & dem\_24m            & $4096\times4096$  & 4326 \\
S. Florida & naip\_latest        & $3546\times4105$  & 2236 \\
Randstad & dem\_open             & $918\times619$    & 4269 \\
Randstad & landcover\_class      & $1949\times1214$  & 4326 \\
Randstad & orthophoto\_cir       & $2448\times1652$  & 4269 \\
Randstad & orthophoto\_rgb       & $2048\times2048$  & 28992 \\
\bottomrule
\end{tabular}
\end{table}

\subsection{Data Sources}

Each area of interest is populated from public data portals: the USGS National Map~\cite{usgs} (elevation and imagery), US Census TIGER/Line files~\cite{tiger} (transportation and administrative boundaries), OpenStreetMap~\cite{osm} (transportation, facilities, and hydrography), regional open-data portals~\cite{nycopen,sfopen,denveropen,harrisopen,colorado,miami,sfwmd} (parcels, flood hazards, and administrative datasets), the FEMA National Flood Hazard Layer~\cite{fema}, and the Dutch PDOK and CBS services~\cite{pdok,cbs} (addresses, neighborhoods, municipalities, elevation, and orthophotography).

\subsection{Derived and Modified Layers}

Most layers are ingested unmodified from their public source. Three categories are derived so that data-quality caveats rest on real conditions rather than hypothetical ones.

\textbf{Controlled geometric defects.} Clean cadastral layers are perturbed by deterministic, seeded transformations to produce layers containing overlapping parcels (a small positive buffer applied to a sampled subset, appended to the layer) and narrow slivers (an inward buffer with the resulting ring retained). Sampling uses fixed strides and fixed random seeds, so the defects are identical on every regeneration. The released tasks exercise three such layers.

\textbf{Derived raster products.} Where no suitable public product exists, a raster is computed from the elevation gradient of the area's digital elevation model and rescaled to a documented range, producing surface-cost and hazard rasters that inherit the source grid exactly.

\textbf{Schema normalization.} Every ingested vector layer is normalized on import, which adds stable identifier fields and, for shapefile outputs, truncates field names to the format's ten-character limit. This is part of geographic recasting rather than incidental: an agent cannot fall back on a field name memorized from the source thread.

\section{Ground Truth Generation and Verification}
\label{app:gt}

\subsection{Generation}

Reference trajectories and their output files are generated and verified with Claude Fable~5, a current frontier model that is deliberately excluded from the set of evaluated agents so that no evaluated model participates in authoring the ground truth it is scored against.

Authoring runs in a deliberately \emph{privileged} setting: the author is given direct access to the input data and to free-form Python for deriving and cross-checking what the correct result should be, neither of which the evaluated agents receive. The released trajectory, however, is expressed entirely in the same 128-API harness the agents are given, and is executed inside that harness to materialize the gold output file. The intended answer is therefore established outside the harness, while the output against which agents are scored is produced inside it.

Reference trajectories are emitted as an ordered list of API calls, each naming an API, its bound arguments, and a handle for its output. The generation prompt constrains the author to the published harness:

\begin{quote}\small\itshape
Use only tools from the provided parsed catalog. Do not assume an unlisted tool, unlisted input field, external dataset, manual edit, hidden preprocessing step, or alternate interpretation of the task.
\end{quote}

This constraint is enforced rather than merely requested: the executor rejects any reference step naming a function outside the set exposed to agents, so a reference that drifted from the harness cannot be executed or released.

\subsection{Automated Verification}

Generated sequences pass through static validation before execution. Argument names are checked against the real API signatures, output handles are checked for binding, and unresolvable references are rejected. Sequences that fail are sent to a verify-and-repair pass, which may rename APIs, correct arguments, or reorder steps, but may not change the task being solved. The verification instruction is quoted verbatim:

\begin{quote}\small\itshape
Check a proposed ground-truth tool sequence for basic tool-use problems.
Check tool names, argument names, placeholders, missing steps, and final output paths.
Do not change the task. If the tool list cannot support the task as written, return reject.
Return exactly one JSON object matching the schema.
\end{quote}

The instruction is deliberately narrow: the verifier is a tool-use checker, not a second author. A sequence that cannot be repaired within the published harness is rejected rather than rewritten, which is why tasks leave the pipeline instead of being weakened to fit it. Surviving sequences are executed in the benchmark environment; a task is retained only if execution yields a well-formed file conforming to its output contract.

\subsection{Expert Review}

Automated checks establish that a trajectory is well formed and runs to completion; they cannot establish that it computes the intended quantity. Reference trajectories were therefore reviewed directly by three GIS experts among the authors, hand-traced step by step, with manual inspection of intermediate outputs and visual inspection of the resulting spatial outputs. Every released reference trajectory passed through this review; a subset was additionally re-verified in a dedicated audit round, for which the per-task records are most complete.

Expert review surfaced defect classes that automated validation cannot detect, because in each case the trajectory executes successfully and produces a plausible output:

\begin{itemize}
\item a field calculation bound to a column that the upstream operation never produced, yielding an all-zero or null reference;
\item a measurement computed on data in geographic coordinates, so that areas and distances are expressed in degree-based units;
\item slope or terrain derivatives computed on an unprojected elevation model;
\item an output contract keyed on a column name the prompt never promised, or renamed during export;
\item off-by-one and truncation errors in row matching;
\item geometry measured before an invalid-geometry repair step rather than after.
\end{itemize}

Tasks whose problem statement proved ambiguous on review, or whose specification would have had to change to become solvable, were removed from the release rather than repaired in place.

\section{Evaluation Metric Definitions}
\label{app:metrics}

This appendix gives the full definitions summarized in Section~3.2 of the main paper. All scoring is deterministic: no language model participates in judging, and no metric compares an agent's trajectory to the reference trajectory.

\subsection{Strict Task Success Rate}

A run is counted correct only when every structural check declared by the task passes and every matched value lies within the task's declared tolerance. Structural checks cover the coordinate reference system, geometry type, data type, raster dimensions and NoData encoding, and the required output columns. Tolerance is applied per task against a combined absolute tolerance of $1e{-}6$ and relative tolerance of $0.01\%$, so scoring is tolerance-aware rather than byte-identical: a result differing only by floating-point noise passes, while one differing by a genuine unit or projection error does not. Strict TSR is binary per task and admits no partial credit.

\subsection{Row-Set Agreement}

Rows are matched as a \emph{set}, never by position. The matcher tries, in order: the contract identifier columns declared by the task; a geometry key, pairing each row with a geometry entity; a detected unique column; and, as a last resort, an optimal assignment over values. The last fallback is optimistic---it finds the most favourable pairing---so closeness computed from those rows should be read as an upper bound. We report precision, recall, and $F_1$ over the resulting correspondence. We refer to recall---the fraction of reference rows matched---as \emph{row coverage}, since it is the quantity that exposes a partial output: a model that emits few rows but gets those rows right scores high on closeness and low on coverage.

\subsection{Non-Spatial Attributes}

Numeric columns are scored as one minus a normalized error, clipped to $[0,1]$. We report symmetric percentage error closeness and root-mean-square error normalized by the reference mean; the underlying record additionally retains weighted mean absolute percentage error, correlation-based agreement, and the fraction of cells within the task's declared tolerance.

Categorical columns are scored with weighted $F_1$, restricted to reference columns carrying more than one class. Single-class columns are excluded because any output that emits the constant value scores perfectly on them, which inflates the aggregate.

Both families of metric are computed over matched rows only. They must therefore be read together with row-set coverage: a model that emits few rows but gets those rows right scores high on closeness and low on coverage, and the coverage-weighted product is the single number that penalizes partial outputs.

\subsection{Spatial Attributes}

Geometry is scored by entity-level matching. For each reference geometry we ask whether a corresponding output geometry exists within tolerance: polygons must reach an intersection-over-union of at least $0.5$, while points and lines must fall within a distance threshold $\varepsilon$. Entity recall is the fraction of reference entities found, entity precision the fraction of produced entities corresponding to a reference entity, and we report entity $F_1$ separately for multi-point, multi-polyline, and multi-polygon outputs.

We prefer entity-level matching to a set-based intersection-over-union for two reasons. First, set-based IoU is only computable when the union of geometries is tractable, so it is evaluated on a self-selected subset of outputs: a model that emits fewer or simpler outputs is scored on a smaller sample, and raw IoU consequently misranks models unless weighted by task coverage. Second, a set-based overlap can look healthy while the number of distinct entities is wrong, which entity-level matching detects and IoU does not.

\subsection{Trajectory Closeness}

Trajectory closeness measures how much of a task's reference workflow an agent actually performed, independently of whether its final output was correct. It is reported as a diagnostic and is not part of either headline metric.

The comparison is made at a coarse functional level rather than over exact API names. Several APIs in the harness accomplish the same operation---two-layer and multi-layer union, for instance, differ only in arity---so matching on exact names would penalize an agent for choosing a legitimate alternative to the one the reference happened to use. Grouping APIs by function first, and then comparing the resulting sets, separates \emph{which operations were performed} from \emph{which particular call was used to perform them}. Read-only inspection calls are excluded on both sides: describing a layer or counting its features advances the workflow no further, so what remains on each side are the calls that change state.

\textbf{Roles.} Each of the 128 harness APIs is mapped to a semantic \emph{role} through a fixed lookup table (128 tools, roughly 103 distinct roles); composite tools map to several roles at once. The mapping is deterministic and applied mechanically, so turning a trajectory into its coarse-grained role sequence involves no model judgment.

The table was built in two independent passes and then reconciled. The first grouped tools by the GIS operation they perform; the second, constructed blind to the first, grouped them by the underlying QGIS or GDAL processing algorithm each one dispatches to. Agreement between the passes was $0.27$ by adjusted Rand index---moderate, as expected for two genuinely different constructs---and all 68 disagreeing tool pairs were individually adjudicated, which surfaced and corrected five real labeling errors in the first pass.

\textbf{Score.} A task's \emph{required roles} are the distinct roles appearing in its own reference trajectory, after discarding read-only inspection calls and deduplicating to a set. An agent's roles are obtained by the same mapping over its real trajectory, discarding calls that returned an error, since a failed call did not accomplish the operation. Writing $R$ for the set of required roles and $A$ for the set of roles the agent performed, trajectory closeness is $|R \cap A| / |R|$, where a required role also counts as satisfied if the agent performed a functionally equivalent substitute for it. Operations beyond those the reference used are recorded but never subtracted, so the measure is a recall over required operations with no precision term.

\textbf{Requirement set.} The reference trajectory is treated as \emph{a} correct solution rather than the only one, and it is a hard ceiling: no role is ever required beyond what the reference itself used. A required role is additionally dropped when at least one model that solved the task under strict TSR did so without invoking that role or a substitute, which prevents the metric from penalizing an agent for skipping a step that is demonstrably not necessary. Dropping is evidence-gated and applied task-wide: 156 of the 349 tasks have at least one droppable role, for 375 role drops in total. Eight tasks lose their entire requirement set this way and are reported as ungradeable rather than scored, on the principle that a ratio with an empty denominator is undefined.

\textbf{Order.} Order violations are checked only for a small set of constraints that admit no counterexample in our data, and harmful insertions only for a single gated case. Both are recorded as informational flags and neither is subtracted from the score, consistent with our practice of scoring leniently wherever a convention is ambiguous.

\textbf{Validation.} Two areas of interest---South Florida and Houston, 68 tasks---were held out entirely while the role map was built. Agreement with strict correctness is comparable on the held-out and calibration splits (row-level point-biserial $r=0.440$ versus $0.420$), indicating the mapping is not fitted to the tasks used to construct it. Pooled over all 349 tasks the row-level correlation is $r=0.424$ and the model-level correlation is $r=0.977$.

\textbf{Limitations.} Requirements are role \emph{sets}, so a reference that applies the same operation to two different layers contributes one required role rather than two, and an agent that applies it once receives full credit. The measure compares operation identity only: it does not inspect arguments, so invoking the right operation on the wrong layer or with the wrong parameters still counts as satisfied. These properties are why we treat trajectory closeness as a diagnostic rather than a scored metric, and why a formal trajectory-level metric over the space of equally valid reference trajectories remains open.

\subsection{Trajectory Diagnostics}

For every run we additionally record the trajectory length (number of API calls issued), the tool call error rate (the fraction of calls returning an error, capturing failures in parameter correctness, input validity, and spatial reference handling), the wall-clock time spent in model inference, and the number of tokens consumed.

\section{Per-Family Timing}
\label{app:timing}

Table~\ref{tab:time_family} reports mean per-task harness time broken down by task family, complementing the aggregate figures in the main paper. Harness time is time spent executing GIS operations inside the evaluation framework; it excludes time spent waiting on the model endpoint, which is reported separately in the main text because it is dominated by vendor and network conditions rather than by task difficulty.

\begin{table}[htbp]
\centering\small
\setlength{\tabcolsep}{3pt}
\caption{Mean per-task harness time (seconds) by task family.}
\label{tab:time_family}
\begin{tabular}{lccccc}
\toprule
\textbf{Model} & \textbf{SJ\&G} & \textbf{SO\&SA} & \textbf{SPA} & \textbf{RS\&IA} & \textbf{TM\&HA} \\
\midrule
Gemini-3.1-Pro   & 143 & 151 & 106 &  99 & 125 \\
Claude-4.6-Opus  & 121 & 120 & 113 &  51 & 107 \\
DeepSeek-V4-Pro  & 156 & 143 & 156 &  73 & 120 \\
GPT-5.4          & 142 & 133 & 142 &  52 &  68 \\
Qwen3.6-27B      & 240 & 175 & 122 &  67 & 126 \\
GPT-OSS-120B     & 116 & 130 & 107 &  85 & 161 \\
\bottomrule
\end{tabular}
\end{table}

\section{Token Consumption by Task Family}
\label{app:tokens}

Table~\ref{tab:token_family} reports mean token consumption per task, overall and broken down by task family, complementing the aggregate figures quoted in Section~3. The raster-heavy families are the most token-hungry for most models, while Spatial Joining \& Geocoding is generally the cheapest.

\begin{table}[htbp]
\centering\small
\setlength{\tabcolsep}{1.5pt}
\caption{Mean tokens consumed per task ($\times 10^{3}$), overall and by task family. Lowest value per column in bold.}
\label{tab:token_family}
\begin{tabular}{lcccccc}
\toprule
\textbf{Model} & \textbf{All} & \textbf{SJ\&G} & \textbf{SO\&SA} & \textbf{SPA} & \textbf{RS\&IA} & \textbf{TM\&HA} \\
\midrule
Gemini-3.1-Pro   & 520 & 407 & 564 & 540 & 657 & 521 \\
Claude-4.6-Opus  & 581 & 495 & 612 & 593 & 612 & 637 \\
DeepSeek-V4-Pro  & 829 & 817 & 807 & 778 & 862 & 898 \\
GPT-5.4          & \textbf{275} & \textbf{256} & \textbf{284} & \textbf{233} & \textbf{314} & \textbf{301} \\
Qwen3.6-27B      & 661 & 716 & 721 & 625 & 576 & 577 \\
GPT-OSS-120B     & 448 & 432 & 474 & 350 & 494 & 483 \\
\bottomrule
\end{tabular}
\end{table}

\section{Case Studies}
\label{app:cases}

Section~3.6 of the main paper states that representative success and failure cases for each task family, with map visualizations of agent and reference outputs, are provided here.

Each case gives the task as the agent received it, the output contract it was scored
against, the reference and agent trajectories, a side-by-side rendering of the two
outputs, and the outcome. Scores are strict pass/fail and quantitative closeness under
the metrics of Appendix~\ref{app:metrics}. Trajectories are reproduced exactly as
executed, with tool names as exposed by the harness.

\subsection{Spatial Joining \& Geocoding}

\subsubsection*{Success: Nearest-destination counting}
\noindent{\small\textit{Task}~\texttt{counting\_\allowbreak destinations\_\allowbreak within\_\allowbreak one\_\allowbreak mile\_\allowbreak of\_\allowbreak origins}}

\noindent\textbf{Objective.} For each Colorado Front Range origin point, count the destination points within 1 mile (1609.344 m) of it.

\noindent\textbf{Constraints.}
\begin{itemize}\small\setlength{\itemsep}{0pt}\setlength{\parskip}{0pt}\setlength{\topsep}{2pt}
\item Search radius is 1 mile = 1609.344 m; destination points exactly at the boundary are included; measure in EPSG:26913.
\item Attach DEST\_CNT\_1 (integer destination count) to each origin point and preserve original origin attributes. Output CRS: EPSG:26913.
\end{itemize}

\noindent\textbf{Output contract.}
\begin{itemize}\small\setlength{\itemsep}{0pt}\setlength{\parskip}{0pt}\setlength{\topsep}{2pt}
\item \texttt{origins\_\allowbreak with\_\allowbreak counts} --- feature class; geometry Point; CRS EPSG:26913; required columns \texttt{DEST\_\allowbreak CNT\_\allowbreak 1}.
\end{itemize}

\noindent\textbf{Reference trajectory} (11 steps).

{\small\raggedright \texttt{reproject\_\allowbreak features} $\to$ \texttt{reproject\_\allowbreak features} $\to$ \texttt{get\_\allowbreak feature\_\allowbreak id\_\allowbreak field} $\to$ \texttt{calculate\_\allowbreak attribute\_\allowbreak field} $\to$ \texttt{calculate\_\allowbreak attribute\_\allowbreak field} $\to$ \texttt{calculate\_\allowbreak attribute\_\allowbreak field} $\to$ \texttt{nearest\_\allowbreak features\_\allowbreak table} $\to$ \texttt{frequency\_\allowbreak table} $\to$ \texttt{join\_\allowbreak table\_\allowbreak by\_\allowbreak field} $\to$ \texttt{calculate\_\allowbreak attribute\_\allowbreak field} $\to$ \texttt{save\_\allowbreak features}\par}

\noindent\textbf{Agent trajectory.} Gemini-3.1-Pro, 9 calls, 0 errors -- reprojects both layers, then collapses the reference's three-step count logic into a single join\_attributes\_by\_location\_within\_distance call, followed by a field rename and cleanup. A shorter, structurally different route to the same answer.

\noindent\textbf{Outcome.} Pass, quantitative closeness 1.00 (all 6 models pass this task unanimously; Gemini-3.1-Pro shown as representative).

\begin{figure}[h]\centering
\includegraphics[width=\linewidth]{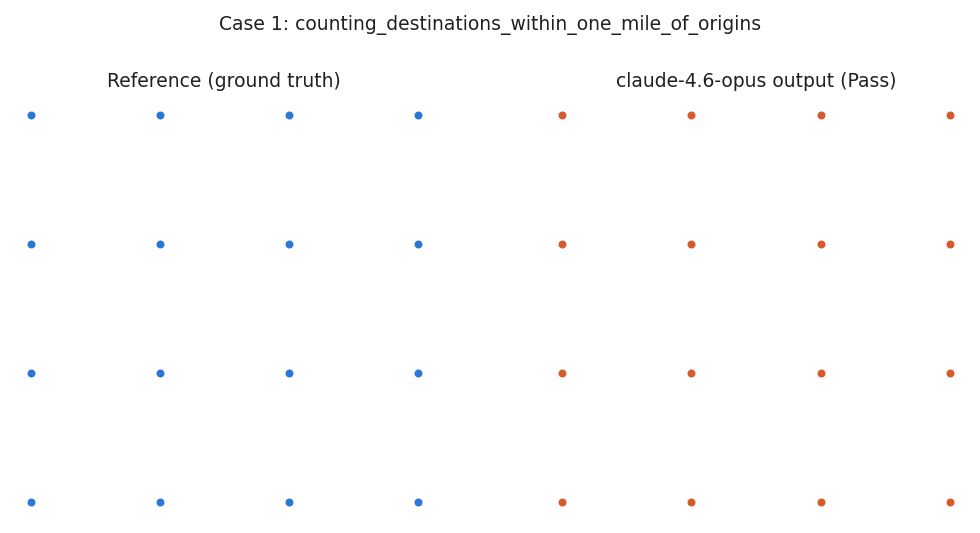}
\caption{Success case, Spatial Joining \& Geocoding: reference output (left) against agent output (right).}
\end{figure}

\subsubsection*{Failure: Interior points from unprojected geometry}
\noindent{\small\textit{Task}~\texttt{creating\_\allowbreak interior\_\allowbreak points\_\allowbreak with\_\allowbreak projected\_\allowbreak coordinates}}

\noindent\textbf{Objective.} For each SF Bay neighborhood polygon, produce one interior point that is guaranteed to lie inside the polygon and attach the polygon's native FID and the point's projected coordinates.

\noindent\textbf{Constraints.}
\begin{itemize}\small\setlength{\itemsep}{0pt}\setlength{\parskip}{0pt}\setlength{\topsep}{2pt}
\item Each output point must lie inside its source polygon (not just on the boundary); use EPSG:26910 for coordinate computation.
\item Output point fields: POLY\_FID (native polygon FID), X\_EPSG2691 (X coordinate in meters, rounded to 2 decimal places), and Y\_EPSG2691 (Y coordinate in meters, rounded to 2 decimal places); preserve original polygon attributes. Output CRS: EPSG:26910.
\end{itemize}

\noindent\textbf{Output contract.}
\begin{itemize}\small\setlength{\itemsep}{0pt}\setlength{\parskip}{0pt}\setlength{\topsep}{2pt}
\item \texttt{artifact\_\allowbreak point\_\allowbreak shapefile} --- feature class; geometry Point; CRS EPSG:26910; required columns \texttt{POLY\_\allowbreak FID}, \texttt{X\_\allowbreak EPSG2691}, \texttt{Y\_\allowbreak EPSG2691}.
\end{itemize}

\noindent\textbf{Reference trajectory} (8 steps).

{\small\raggedright \texttt{reproject\_\allowbreak features} $\to$ \texttt{get\_\allowbreak feature\_\allowbreak id\_\allowbreak field} $\to$ \texttt{calculate\_\allowbreak attribute\_\allowbreak field} $\to$ \texttt{point\_\allowbreak on\_\allowbreak surface} $\to$ \texttt{add\_\allowbreak geometry\_\allowbreak attributes} $\to$ \texttt{calculate\_\allowbreak attribute\_\allowbreak field} $\to$ \texttt{calculate\_\allowbreak attribute\_\allowbreak field} $\to$ \texttt{save\_\allowbreak features}\par}

\noindent\textbf{Agent trajectory.} DeepSeek-V4-Pro, 14 calls, 0 errors -- calls point\_on\_surface before reprojecting, then reprojects afterward and derives coordinates with add\_xy\_fields instead of add\_geometry\_attributes. Divergence point: step 1. Computing the interior point before projecting yields a different point than computing it after --- both points fall inside the same poly

\noindent\textbf{Outcome.} Fail, quantitative closeness 0.00, 0 tool-call errors --- the run completes cleanly and produces a valid-looking point layer that is simply the wrong point.

\begin{figure}[h]\centering
\includegraphics[width=\linewidth]{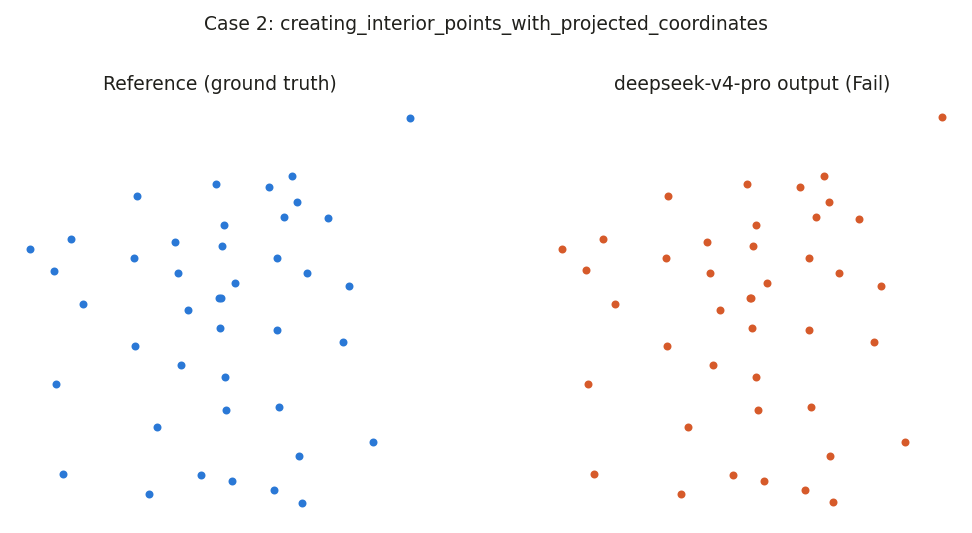}
\caption{Failure case, Spatial Joining \& Geocoding: reference output (left) against agent output (right).}
\end{figure}

\subsection{Spatial Overlay \& Suitability Analysis}

\subsubsection*{Success: Inward county boundary inset}
\noindent{\small\textit{Task}~\texttt{creating\_\allowbreak inland\_\allowbreak county\_\allowbreak boundary\_\allowbreak inset}}

\noindent\textbf{Objective.} From the Harris County boundary polygon, produce the polygon representing an inward inset of 3,000 m.

\noindent\textbf{Constraints.}
\begin{itemize}\small\setlength{\itemsep}{0pt}\setlength{\parskip}{0pt}\setlength{\topsep}{2pt}
\item Inset the county polygon boundary 3,000 m inward; keep only valid non-empty polygon parts; each resulting polygon part is a separate output feature.
\item Output fields: orig\_fid (native county polygon FID) and inset\_m (= 3000 for all features). Output CRS: EPSG:26915.
\end{itemize}

\noindent\textbf{Output contract.}
\begin{itemize}\small\setlength{\itemsep}{0pt}\setlength{\parskip}{0pt}\setlength{\topsep}{2pt}
\item \texttt{houston\_\allowbreak county\_\allowbreak inland\_\allowbreak agent} --- feature class; geometry Polygon; CRS EPSG:26915; required columns \texttt{orig\_\allowbreak fid}, \texttt{inset\_\allowbreak m}.
\end{itemize}

\noindent\textbf{Reference trajectory} (10 steps).

{\small\raggedright \texttt{save\_\allowbreak features} $\to$ \texttt{get\_\allowbreak feature\_\allowbreak id\_\allowbreak field} $\to$ \texttt{calculate\_\allowbreak attribute\_\allowbreak field} $\to$ \texttt{reproject\_\allowbreak features} $\to$ \texttt{buffer\_\allowbreak features} $\to$ \texttt{fix\_\allowbreak geometries} $\to$ \texttt{multipart\_\allowbreak to\_\allowbreak singleparts} $\to$ \texttt{calculate\_\allowbreak attribute\_\allowbreak field} $\to$ \texttt{add\_\allowbreak geometry\_\allowbreak attributes} $\to$ \texttt{extract\_\allowbreak features\_\allowbreak by\_\allowbreak expression}\par}

\noindent\textbf{Agent trajectory.} Claude-4.6-Opus, 8 calls, 0 errors -- reproject to buffer\_features to multipart\_to\_singleparts to compute columns to retain\_fields --- omits the reference's fix\_geometries and final expression-based cleanup. A shorter path that happens not to need the cleanup steps on this particular input, since the source boundary has no pre-existing topology

\noindent\textbf{Outcome.} Pass, quantitative closeness 1.00 (all 6 models pass unanimously; Claude-4.6-Opus shown as representative).

\begin{figure}[h]\centering
\includegraphics[width=\linewidth]{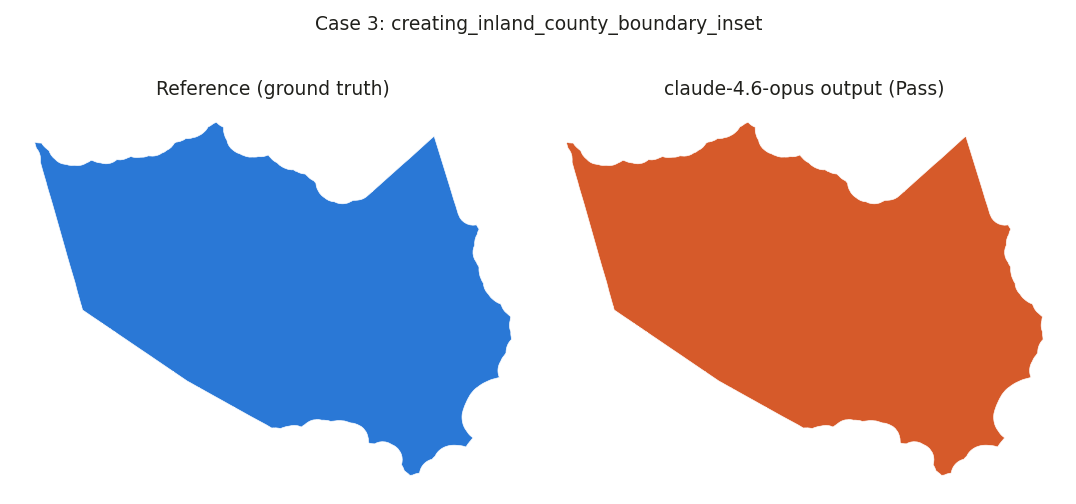}
\caption{Success case, Spatial Overlay \& Suitability Analysis: reference output (left) against agent output (right).}
\end{figure}

\subsubsection*{Failure: Shoreline clipping with a CRS-preservation trap}
\noindent{\small\textit{Task}~\texttt{clipping\_\allowbreak shoreline\_\allowbreak to\_\allowbreak boroughs}}

\noindent\textbf{Objective.} Clip the NYC shoreline lines to the borough polygons and report the clipped length of each original shoreline segment.

\noindent\textbf{Constraints.}
\begin{itemize}\small\setlength{\itemsep}{0pt}\setlength{\parskip}{0pt}\setlength{\topsep}{2pt}
\item Retain only shoreline geometry that falls inside borough polygons; measure lengths in EPSG:2263.
\item Output: clipped shoreline lines in the input shoreline CRS, and a CSV with columns: source\_fid (native shoreline FID) and length\_m (clipped length in meters, rounded to 2 decimal places).
\end{itemize}

\noindent\textbf{Output contract.}
\begin{itemize}\small\setlength{\itemsep}{0pt}\setlength{\parskip}{0pt}\setlength{\topsep}{2pt}
\item \texttt{clipped\_\allowbreak shoreline\_\allowbreak agent} --- feature class; geometry Polyline; CRS Preserve the CRS of {input\_dir}/nyc\_new\_york/transport/shoreline.shp (output must use the same CRS).
\item \texttt{clipped\_\allowbreak lengths\_\allowbreak csv} --- csv; required columns \texttt{source\_\allowbreak fid}, \texttt{length\_\allowbreak m}.
\end{itemize}

\noindent\textbf{Reference trajectory} (10 steps).

{\small\raggedright \texttt{reproject\_\allowbreak features} $\to$ \texttt{reproject\_\allowbreak features} $\to$ \texttt{fix\_\allowbreak geometries} $\to$ \texttt{fix\_\allowbreak geometries} $\to$ \texttt{clip\_\allowbreak features} $\to$ \texttt{get\_\allowbreak feature\_\allowbreak id\_\allowbreak field} $\to$ \texttt{calculate\_\allowbreak attribute\_\allowbreak field} $\to$ \texttt{calculate\_\allowbreak attribute\_\allowbreak field} $\to$ \texttt{reproject\_\allowbreak features} $\to$ \texttt{export\_\allowbreak table\_\allowbreak to\_\allowbreak csv}\par}

\noindent\textbf{Agent trajectory.} Gemini-3.1-Pro, 34 calls, 1 tool-call error -- attempts a field calculation before reprojecting, then reprojects and clips --- this first clip\_features call fails with a real tool error (TopologyException: side location conflict, an invalid-geometry symptom) because fix\_geometries had not yet been run. The agent recovers by fixing geometries and re-clipping, then spend

\noindent\textbf{Outcome.} Fail, quantitative closeness 0.499 --- among the closest near-misses in this appendix; only DeepSeek-V4-Pro (1/6) preserves the input CRS correctly on this task.

\begin{figure}[h]\centering
\includegraphics[width=\linewidth]{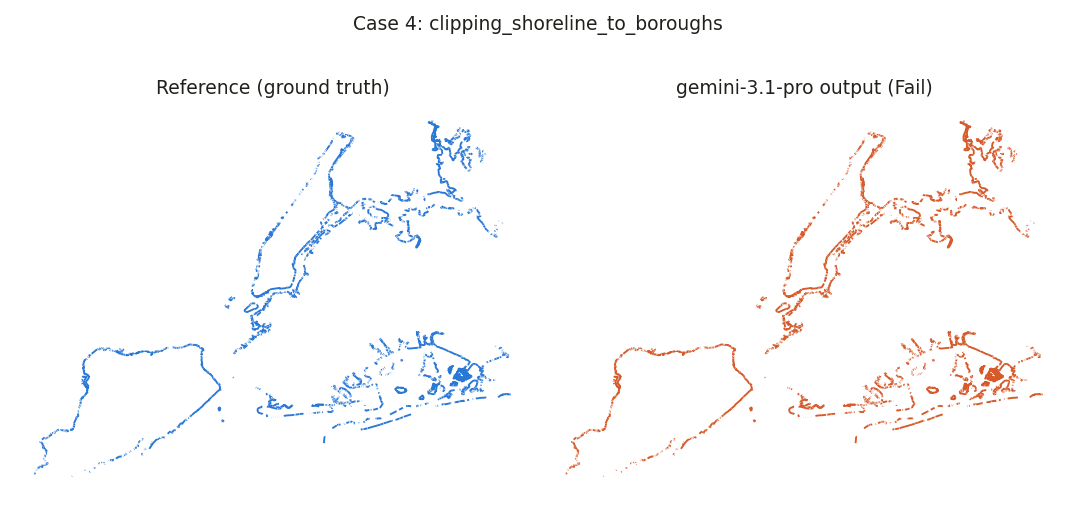}
\caption{Failure case, Spatial Overlay \& Suitability Analysis: reference output (left) against agent output (right).}
\end{figure}

\subsection{Terrain Modeling \& Hydrological Analysis}

\subsubsection*{Success: Canal network topology classification}
\noindent{\small\textit{Task}~\texttt{classifying\_\allowbreak multichannel\_\allowbreak canal\_\allowbreak segments}}

\noindent\textbf{Objective.} From the Houston canal line network, classify each planarized segment as multichannel or single-channel and report its length.

\noindent\textbf{Constraints.}
\begin{itemize}\small\setlength{\itemsep}{0pt}\setlength{\parskip}{0pt}\setlength{\topsep}{2pt}
\item Planarize the network by splitting lines at all intersections; an 'island polygon' is any positive-area polygon enclosed by the planarized linework.
\item A segment is multichannel (is\_multich = 1) when it intersects any island polygon; otherwise it is single-channel (is\_multich = 0).
\item Output fields: ORIG\_FID (native source line FID), is\_multich (0 or 1), and length\_m (segment length in meters, measured in EPSG:26915, rounded to 2 decimal places); output CRS: EPSG:26915.
\end{itemize}

\noindent\textbf{Output contract.}
\begin{itemize}\small\setlength{\itemsep}{0pt}\setlength{\parskip}{0pt}\setlength{\topsep}{2pt}
\item \texttt{classified\_\allowbreak canals} --- feature class; geometry Polyline; CRS EPSG:26915; required columns \texttt{ORIG\_\allowbreak FID}, \texttt{is\_\allowbreak multich}, \texttt{length\_\allowbreak m}.
\end{itemize}

\noindent\textbf{Reference trajectory} (12 steps).

{\small\raggedright \texttt{reproject\_\allowbreak features} $\to$ \texttt{get\_\allowbreak feature\_\allowbreak id\_\allowbreak field} $\to$ \texttt{calculate\_\allowbreak attribute\_\allowbreak field} $\to$ \texttt{features\_\allowbreak to\_\allowbreak lines} $\to$ \texttt{polygonize\_\allowbreak lines} $\to$ \texttt{get\_\allowbreak feature\_\allowbreak id\_\allowbreak field} $\to$ \texttt{calculate\_\allowbreak attribute\_\allowbreak field} $\to$ \texttt{join\_\allowbreak attributes\_\allowbreak by\_\allowbreak location\_\allowbreak summary} $\to$ \texttt{calculate\_\allowbreak attribute\_\allowbreak field} $\to$ \texttt{calculate\_\allowbreak attribute\_\allowbreak field} $\to$ \texttt{retain\_\allowbreak fields} $\to$ \texttt{save\_\allowbreak features}\par}

\noindent\textbf{Agent trajectory.} Claude-4.6-Opus, 14 calls, 0 errors -- follows the same planarization (features\_to\_lines to polygonize\_lines), then substitutes a plain join\_attributes\_by\_location plus its own add\_geometry\_attributes / extract\_features\_by\_expression steps for the reference's single aggregating join\_attributes\_by\_location\_summary call --- a mo

\noindent\textbf{Outcome.} Pass, quantitative closeness 1.00 (all 6 models pass unanimously --- the only Terrain Modeling task in this benchmark with a clean sweep; Claude-4.6-Opus shown as representative).

\begin{figure}[h]\centering
\includegraphics[width=\linewidth]{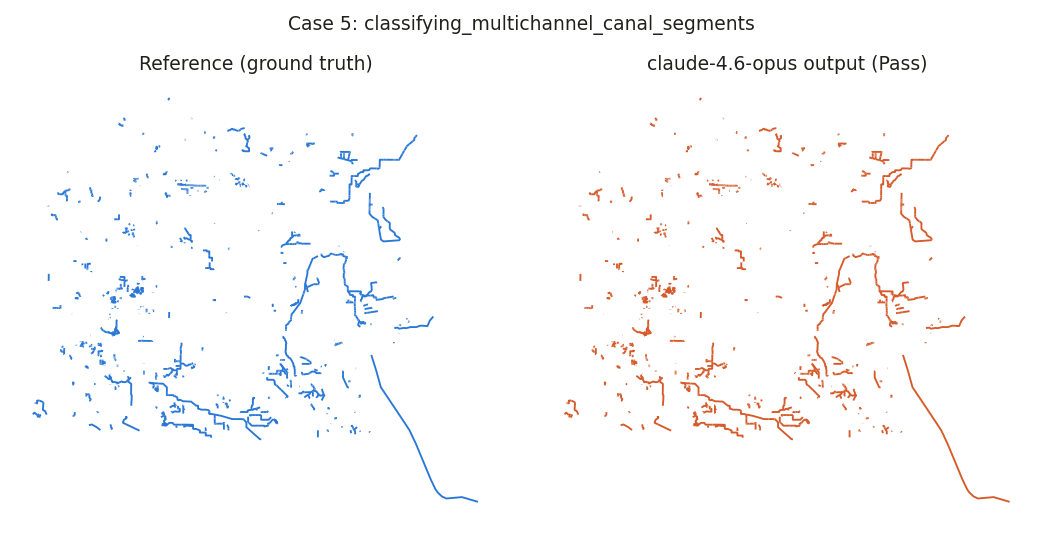}
\caption{Success case, Terrain Modeling \& Hydrological Analysis: reference output (left) against agent output (right).}
\end{figure}

\subsubsection*{Failure: Steep-slope polygonization}
\noindent{\small\textit{Task}~\texttt{polygonizing\_\allowbreak steep\_\allowbreak slope\_\allowbreak areas}}

\noindent\textbf{Objective.} From the Colorado Front Range DEM, produce polygons for contiguous areas where slope (on a 10 m grid) is at least 20 percent.

\noindent\textbf{Constraints.}
\begin{itemize}\small\setlength{\itemsep}{0pt}\setlength{\parskip}{0pt}\setlength{\topsep}{2pt}
\item Slope in percent at 10 m analysis resolution; cells with slope >= 20\% are steep; group contiguous steep cells.
\item Output polygon fields: raster\_value (class identifier), slope\_\_pct (maximum slope percent within the polygon, rounded to 2 decimal places), and area\_m2 (area in m2, measured in EPSG:26913, rounded to 2 decimal places); output CRS: EPSG:26913.
\end{itemize}

\noindent\textbf{Output contract.}
\begin{itemize}\small\setlength{\itemsep}{0pt}\setlength{\parskip}{0pt}\setlength{\topsep}{2pt}
\item \texttt{agent\_\allowbreak shapefile} --- feature class; geometry Polygon; CRS EPSG:26913; required columns \texttt{raster\_\allowbreak val}, \texttt{slope\_\allowbreak \_\allowbreak pct}, \texttt{area\_\allowbreak m2}.
\end{itemize}

\noindent\textbf{Reference trajectory} (17 steps).

{\small\raggedright \texttt{warp\_\allowbreak reproject\_\allowbreak raster} $\to$ \texttt{slope\_\allowbreak raster} $\to$ \texttt{raster\_\allowbreak calculator} $\to$ \texttt{raster\_\allowbreak calculator} $\to$ \texttt{polygonize\_\allowbreak raster} $\to$ \texttt{extract\_\allowbreak features\_\allowbreak by\_\allowbreak expression} $\to$ \texttt{reproject\_\allowbreak features} $\to$ \texttt{get\_\allowbreak feature\_\allowbreak id\_\allowbreak field} $\to$ \texttt{calculate\_\allowbreak attribute\_\allowbreak field} $\to$ \texttt{zonal\_\allowbreak statistics\_\allowbreak as\_\allowbreak table} $\to$ \texttt{join\_\allowbreak table\_\allowbreak by\_\allowbreak field} $\to$ \texttt{calculate\_\allowbreak attribute\_\allowbreak field} $\to$ \texttt{calculate\_\allowbreak attribute\_\allowbreak field} $\to$ \texttt{add\_\allowbreak geometry\_\allowbreak attributes} $\to$ \texttt{calculate\_\allowbreak attribute\_\allowbreak field} $\to$ \texttt{retain\_\allowbreak fields} $\to$ \texttt{save\_\allowbreak features}\par}

\noindent\textbf{Agent trajectory.} Claude-4.6-Opus, 19 calls, 1 tool-call error -- matches the reference up through slope thresholding, but inserts an extra clump\_raster\_regions call before polygonize\_raster to pre-group contiguous cells --- redundant with what polygonize\_raster already does on its own. Divergence point: this inserted step. It changes the downstream file/ID chain enoug

\noindent\textbf{Outcome.} Fail, quantitative closeness 0.00 (0/6 models pass; four of six complete with 0--1 tool errors, so the miss is in the final polygon set, not the process).

\begin{figure}[h]\centering
\includegraphics[width=\linewidth]{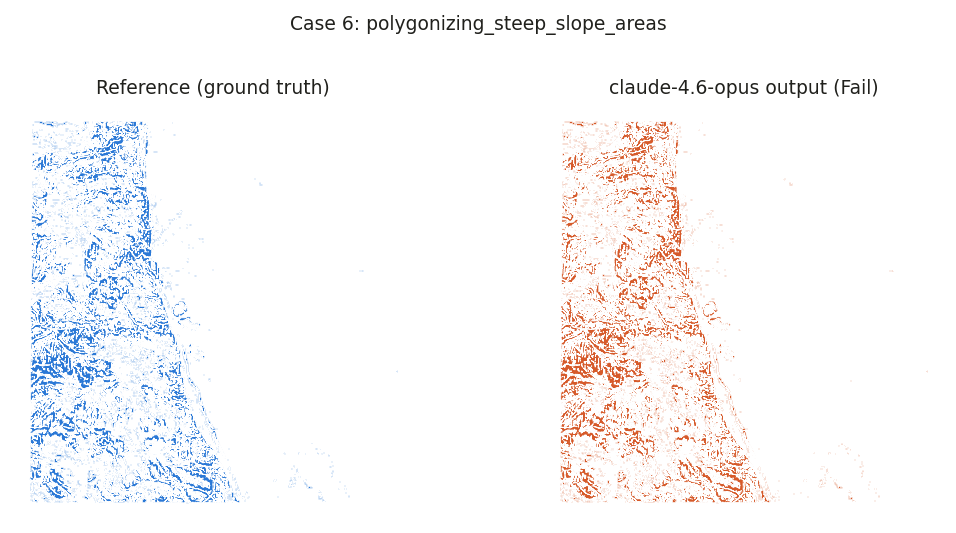}
\caption{Failure case, Terrain Modeling \& Hydrological Analysis: reference output (left) against agent output (right).}
\end{figure}

\subsection{Spatial Pattern Analysis}

\subsubsection*{Success: Park-overlap grid counting}
\noindent{\small\textit{Task}~\texttt{counting\_\allowbreak park\_\allowbreak overlaps\_\allowbreak in\_\allowbreak grid\_\allowbreak cells}}

\noindent\textbf{Objective.} Over the South Florida park extent, create a 1 km polygon grid and count how many distinct park polygons intersect each cell.

\noindent\textbf{Constraints.}
\begin{itemize}\small\setlength{\itemsep}{0pt}\setlength{\parskip}{0pt}\setlength{\topsep}{2pt}
\item Grid cell size: 1,000 m; each input park feature contributes at most 1 to a cell count regardless of how many polygons it has; cells with no intersecting parks have park\_count = 0.
\item Output grid polygon fields: grid\_id (integer cell identifier) and park\_count; output CRS: EPSG:32617.
\end{itemize}

\noindent\textbf{Output contract.}
\begin{itemize}\small\setlength{\itemsep}{0pt}\setlength{\parskip}{0pt}\setlength{\topsep}{2pt}
\item \texttt{richness\_\allowbreak grid} --- feature class; geometry Polygon; CRS EPSG:32617; required columns \texttt{grid\_\allowbreak id}, \texttt{park\_\allowbreak count}.
\end{itemize}

\noindent\textbf{Reference trajectory} (11 steps).

{\small\raggedright \texttt{reproject\_\allowbreak features} $\to$ \texttt{get\_\allowbreak feature\_\allowbreak id\_\allowbreak field} $\to$ \texttt{calculate\_\allowbreak attribute\_\allowbreak field} $\to$ \texttt{fix\_\allowbreak geometries} $\to$ \texttt{indexed\_\allowbreak fishnet} $\to$ \texttt{intersection\_\allowbreak overlay} $\to$ \texttt{frequency\_\allowbreak table} $\to$ \texttt{join\_\allowbreak table\_\allowbreak by\_\allowbreak field} $\to$ \texttt{calculate\_\allowbreak attribute\_\allowbreak field} $\to$ \texttt{save\_\allowbreak features} $\to$ \texttt{retain\_\allowbreak fields}\par}

\noindent\textbf{Agent trajectory.} Claude-4.6-Opus, 10 calls, 0 errors -- builds the same grid, then collapses the reference's three-step overlay-count-join sequence into a single join\_attributes\_by\_location\_summary call --- the same kind of aggregating-join shortcut seen in the canal-classification success case above.

\noindent\textbf{Outcome.} Pass, quantitative closeness 1.00 (4/6 models pass; Claude-4.6-Opus shown as representative).

\begin{figure}[h]\centering
\includegraphics[width=\linewidth]{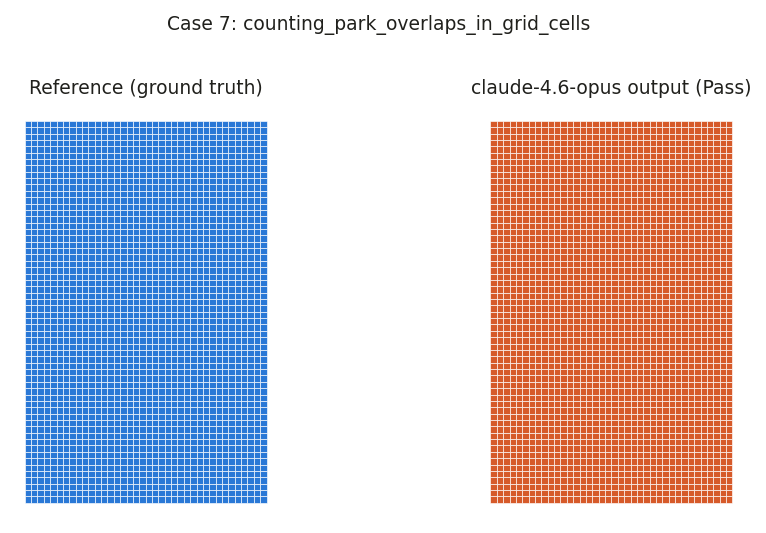}
\caption{Success case, Spatial Pattern Analysis: reference output (left) against agent output (right).}
\end{figure}

\subsubsection*{Failure: Voronoi-clipped hospital buffers}
\noindent{\small\textit{Task}~\texttt{sf\_\allowbreak bay\_\allowbreak voronoi\_\allowbreak clipped\_\allowbreak hospital\_\allowbreak buffers\_\allowbreak 1\_\allowbreak km}}

\noindent\textbf{Objective.} For each SF Bay hospital, produce a polygon representing the intersection of a 1,000 m circle around it and its bounded Voronoi cell within the neighborhood boundaries.

\noindent\textbf{Constraints.}
\begin{itemize}\small\setlength{\itemsep}{0pt}\setlength{\parskip}{0pt}\setlength{\topsep}{2pt}
\item Voronoi cells are derived from hospital points, bounded by the combined neighborhood extent; each hospital's output polygon is its 1,000 m circle intersected with its Voronoi cell; measure in EPSG:26910.
\item Output polygon fields: hospit\_fid (native hospital FID) and buffer\_rad (= 1000); output CRS: EPSG:26910.
\end{itemize}

\noindent\textbf{Output contract.}
\begin{itemize}\small\setlength{\itemsep}{0pt}\setlength{\parskip}{0pt}\setlength{\topsep}{2pt}
\item \texttt{agent\_\allowbreak shp} --- feature class; geometry Polygon; CRS EPSG:26910; required columns \texttt{hospit\_\allowbreak fid}, \texttt{buffer\_\allowbreak rad}.
\end{itemize}

\noindent\textbf{Reference trajectory} (15 steps).

{\small\raggedright \texttt{reproject\_\allowbreak features} $\to$ \texttt{reproject\_\allowbreak features} $\to$ \texttt{get\_\allowbreak feature\_\allowbreak id\_\allowbreak field} $\to$ \texttt{calculate\_\allowbreak attribute\_\allowbreak field} $\to$ \texttt{voronoi\_\allowbreak polygons} $\to$ \texttt{dissolve\_\allowbreak features} $\to$ \texttt{clip\_\allowbreak features} $\to$ \texttt{buffer\_\allowbreak features} $\to$ \texttt{fix\_\allowbreak geometries} $\to$ \texttt{fix\_\allowbreak geometries} $\to$ \texttt{intersection\_\allowbreak overlay} $\to$ \texttt{extract\_\allowbreak features\_\allowbreak by\_\allowbreak expression} $\to$ \texttt{calculate\_\allowbreak attribute\_\allowbreak field} $\to$ \texttt{retain\_\allowbreak fields} $\to$ \texttt{save\_\allowbreak features}\par}

\noindent\textbf{Agent trajectory.} Gemini-3.1-Pro, 9 calls, 0 errors -- reproject, dissolve, Voronoi, clip --- then substitutes a single specialized clip\_point\_buffers\_to\_containing\_polygons call for the reference's buffer/intersect/expression-filter combination, and skips both fix\_geometries passes entirely. Divergence point: this substitution. The composite tool does not reproduce the reference's

\noindent\textbf{Outcome.} Fail, quantitative closeness 0.00, 0 tool-call errors --- only Claude-4.6-Opus (1/6) solves this task; three challenge dimensions stacked at once make it one of the harder cases in the benchmark. This is also the Spatial Pattern Analysis exemplar shown in Figure~1 of the main paper.

\begin{figure}[h]\centering
\includegraphics[width=\linewidth]{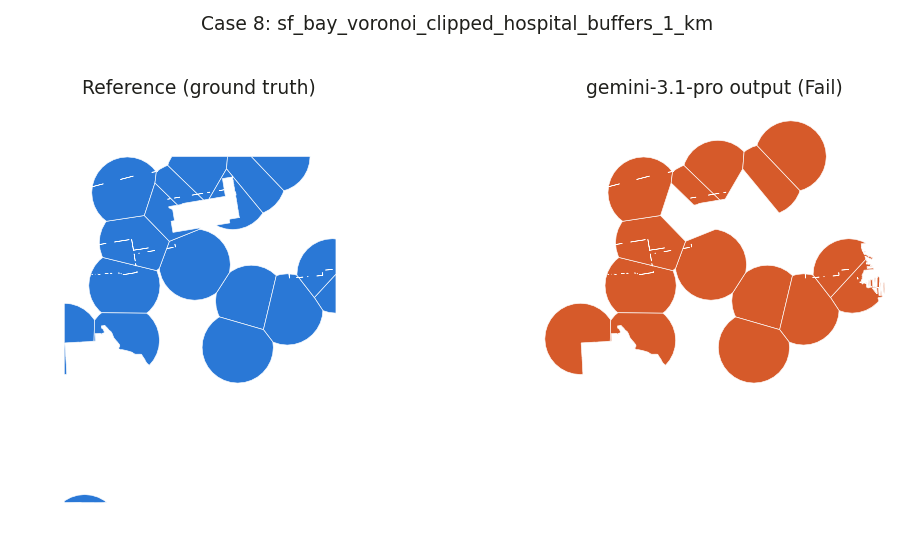}
\caption{Failure case, Spatial Pattern Analysis: reference output (left) against agent output (right).}
\end{figure}

\subsection{Remote Sensing \& Image Analysis}

\subsubsection*{Success: NDVI from a color-infrared orthophoto}
\noindent{\small\textit{Task}~\texttt{extract\_\allowbreak ndvi\_\allowbreak values\_\allowbreak at\_\allowbreak randstad\_\allowbreak address\_\allowbreak points\_\allowbreak from\_\allowbreak cir\_\allowbreak orthophoto}}

\noindent\textbf{Objective.} For each Randstad address point, compute NDVI from the CIR orthophoto at the point location and attach it to the point layer.

\noindent\textbf{Constraints.}
\begin{itemize}\small\setlength{\itemsep}{0pt}\setlength{\parskip}{0pt}\setlength{\topsep}{2pt}
\item NDVI (NIR band 1, R band 2); set NDVI to null when either sampled band is NoData or when NIR + R = 0.
\item NDVI field: Float32, rounded to 4 decimal places; preserve all original address point attributes. Output CRS: EPSG:28992.
\end{itemize}

\noindent\textbf{Output contract.}
\begin{itemize}\small\setlength{\itemsep}{0pt}\setlength{\parskip}{0pt}\setlength{\topsep}{2pt}
\item \texttt{agent\_\allowbreak shapefile} --- feature class; geometry Point; CRS EPSG:28992; required columns \texttt{NDVI}.
\end{itemize}

\noindent\textbf{Reference trajectory} (9 steps).

{\small\raggedright \texttt{reproject\_\allowbreak features} $\to$ \texttt{rearrange\_\allowbreak raster\_\allowbreak bands} $\to$ \texttt{rearrange\_\allowbreak raster\_\allowbreak bands} $\to$ \texttt{raster\_\allowbreak calculator} $\to$ \texttt{sample\_\allowbreak raster\_\allowbreak values\_\allowbreak to\_\allowbreak points} $\to$ \texttt{rename\_\allowbreak field} $\to$ \texttt{calculate\_\allowbreak attribute\_\allowbreak field} $\to$ \texttt{delete\_\allowbreak columns} $\to$ \texttt{save\_\allowbreak features}\par}

\noindent\textbf{Agent trajectory.} Claude-4.6-Opus, 8 calls, 0 errors -- samples the raw orthophoto bands directly at the points first, then computes NDVI as an attribute-field formula on the sampled band values, reprojecting only at the very end. A different order of operations --- computing the index after sampling rather than before --- that lands on the same NDVI values for this raster.

\noindent\textbf{Outcome.} Pass, quantitative closeness 1.00 (3/6 models pass this task; Claude-4.6-Opus shown as representative).

\begin{figure}[h]\centering
\includegraphics[width=\linewidth]{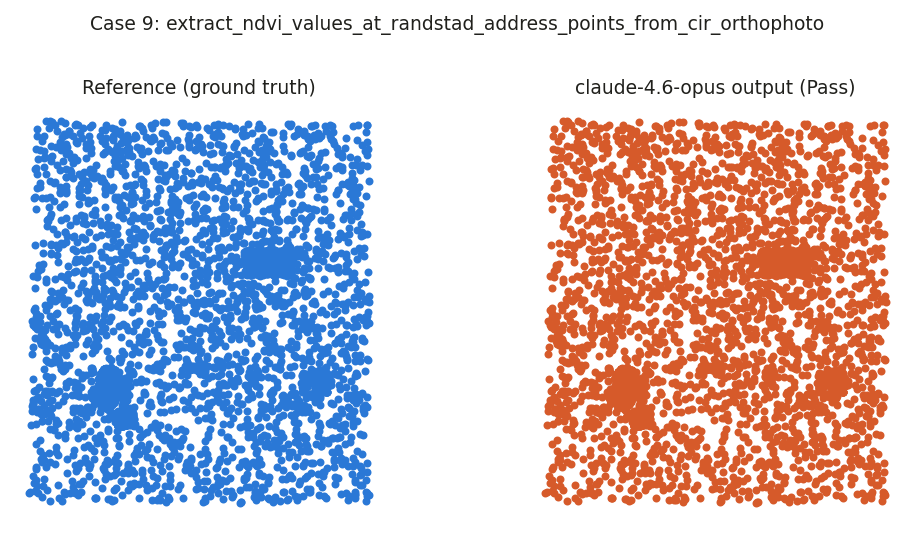}
\caption{Success case, Remote Sensing \& Image Analysis: reference output (left) against agent output (right).}
\end{figure}

\subsubsection*{Failure: Landcover raster polygonization with a skipped reprojection}
\noindent{\small\textit{Task}~\texttt{vectorize\_\allowbreak randstad\_\allowbreak landcover\_\allowbreak raster\_\allowbreak classes\_\allowbreak into\_\allowbreak polygons}}

\noindent\textbf{Objective.} Convert the Randstad landcover raster into vector polygons, grouping 8-connected cells of the same value.

\noindent\textbf{Constraints.}
\begin{itemize}\small\setlength{\itemsep}{0pt}\setlength{\parskip}{0pt}\setlength{\topsep}{2pt}
\item 8-connected regions (including diagonals); one polygon per connected region; raster\_val field stores the integer pixel value.
\item Output CRS: EPSG:28992.
\end{itemize}

\noindent\textbf{Output contract.}
\begin{itemize}\small\setlength{\itemsep}{0pt}\setlength{\parskip}{0pt}\setlength{\topsep}{2pt}
\item \texttt{vector\_\allowbreak shapefile} --- feature class; geometry Polygon; CRS EPSG:28992; required columns \texttt{raster\_\allowbreak val}.
\end{itemize}

\noindent\textbf{Reference trajectory} (5 steps).

{\small\raggedright \texttt{warp\_\allowbreak reproject\_\allowbreak raster} $\to$ \texttt{polygonize\_\allowbreak raster} $\to$ \texttt{fix\_\allowbreak geometries} $\to$ \texttt{calculate\_\allowbreak attribute\_\allowbreak field} $\to$ \texttt{save\_\allowbreak features}\par}

\noindent\textbf{Agent trajectory.} Claude-4.6-Opus, 1 call, 0 errors -- calls polygonize\_raster directly on the raw raster and submits the result. Divergence point: step 1. Skipping warp\_reproject\_raster means the output inherits the source raster's native geographic CRS (EPSG:4326, degrees) instead of the required EPSG:28992 (meters) --- confirmed directly from the two files' CRS metadata. Once reproj

\noindent\textbf{Outcome.} Fail, quantitative closeness 0.00 (0/6 models pass this no-caveat control task; every model completes with zero or one tool-call error, so failure here reflects a specific missed step rather than general raster-to-vector difficulty).

\begin{figure}[h]\centering
\includegraphics[width=\linewidth]{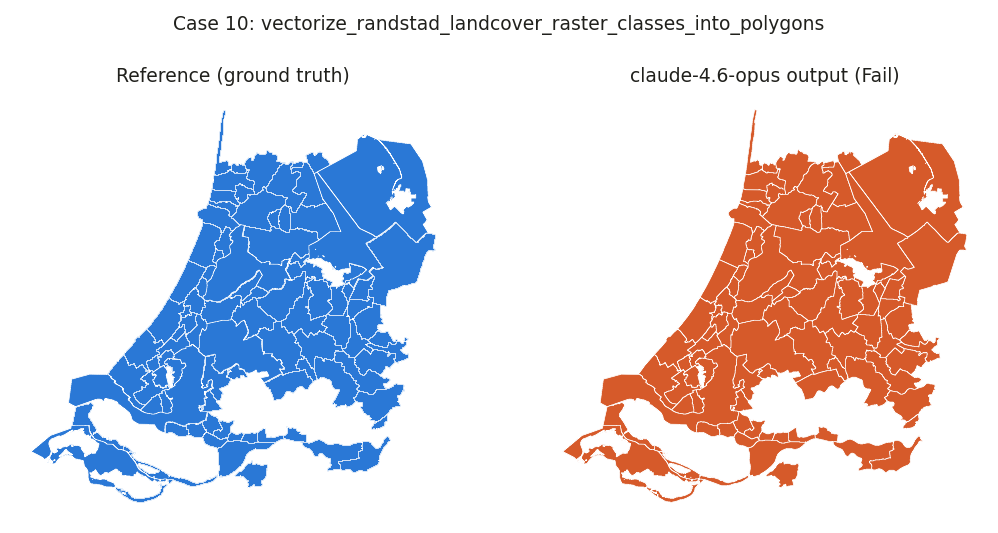}
\caption{Failure case, Remote Sensing \& Image Analysis: reference output (left) against agent output (right).}
\end{figure}

\end{document}